%% file: Thesis.tex
\documentclass[a4paper, 11pt, oneside]{Thesis}  
\graphicspath{Figures/}  

\usepackage[square, numbers, comma, sort&compress]{natbib}  
\usepackage{verbatim}  
\usepackage{vector}  
\hypersetup{urlcolor=blue, colorlinks=true}  

\makeatother

\usepackage{algorithmicx, algorithm, algpseudocode}
\usepackage{amsmath}
\usepackage{amssymb}
\usepackage{wasysym}
\usepackage{setspace}
\usepackage{color}
\usepackage{booktabs}
\usepackage{microtype}
\usepackage{tikz-dependency}
\usepackage{hhline}
\usepackage[colorinlistoftodos]{todonotes}
\usepackage{titlesec}
\usepackage{multirow}
\usepackage{hyperref}
\usepackage{url}
\usepackage{multibib}
\usepackage{bm}
\usepackage[utf8]{inputenc}
\usepackage[T1]{fontenc}
\usepackage[T1]{fontenc}
\usepackage{textcomp}
\usepackage{CJKutf8}
\usepackage{blindtext}
\usepackage[final]{pdfpages}

\algrenewcommand\alglinenumber[1]{{\sffamily\footnotesize#1}}
\newenvironment{nospaceflalign*}
 {\setlength{\abovedisplayskip}{0pt}\setlength{\belowdisplayskip}{0pt}%
  \csname flalign*\endcsname}
 {\csname endflalign*\endcsname\ignorespacesafterend}
 
\newcommand{\smallpar}[1]{\vspace{-3pt}\textbf{#1}}

\newcommand{\smallsubsection}[1]{\vspace{-6pt}\subsection{#1}}
\newcommand{\norm}[1]{\left\lVert#1\right\rVert}

\newcommand{\eat}[1]{\ignorespaces}
\newcommand{\ignore}[1]{}

\newcommand{\eg}{{\sl e.g.}}

\newcommand{\ie}{{\sl i.e.}}

\newcommand{\hhide}[1]{}
\newcommand{\dam}[1]{{\color{red} dam: #1}}

\begin{document}
\begin{CJK*}{UTF8}{gkai}

\frontmatter      

\title  {Knowledge Efficient Deep Learning for Natural Language Processing}
\authors  {\texorpdfstring
            {\href{https://ttic.uchicago.edu/~haiwang/}{Hai Wang}}
            {Hai Wang}
            }
\addresses  {\univname}  
\date       {\today}
\subject    {}
\maketitle
\setstretch{1.3}  

\fancyhead{}  
\rhead{\thepage}  
\lhead{}  

\pagestyle{fancy}  


\ignore{
\Declaration{
\input{Declaration/Declaration.tex}
}
\clearpage
}

\includepdf[pages=-, offset=70 0, scale=1.1]{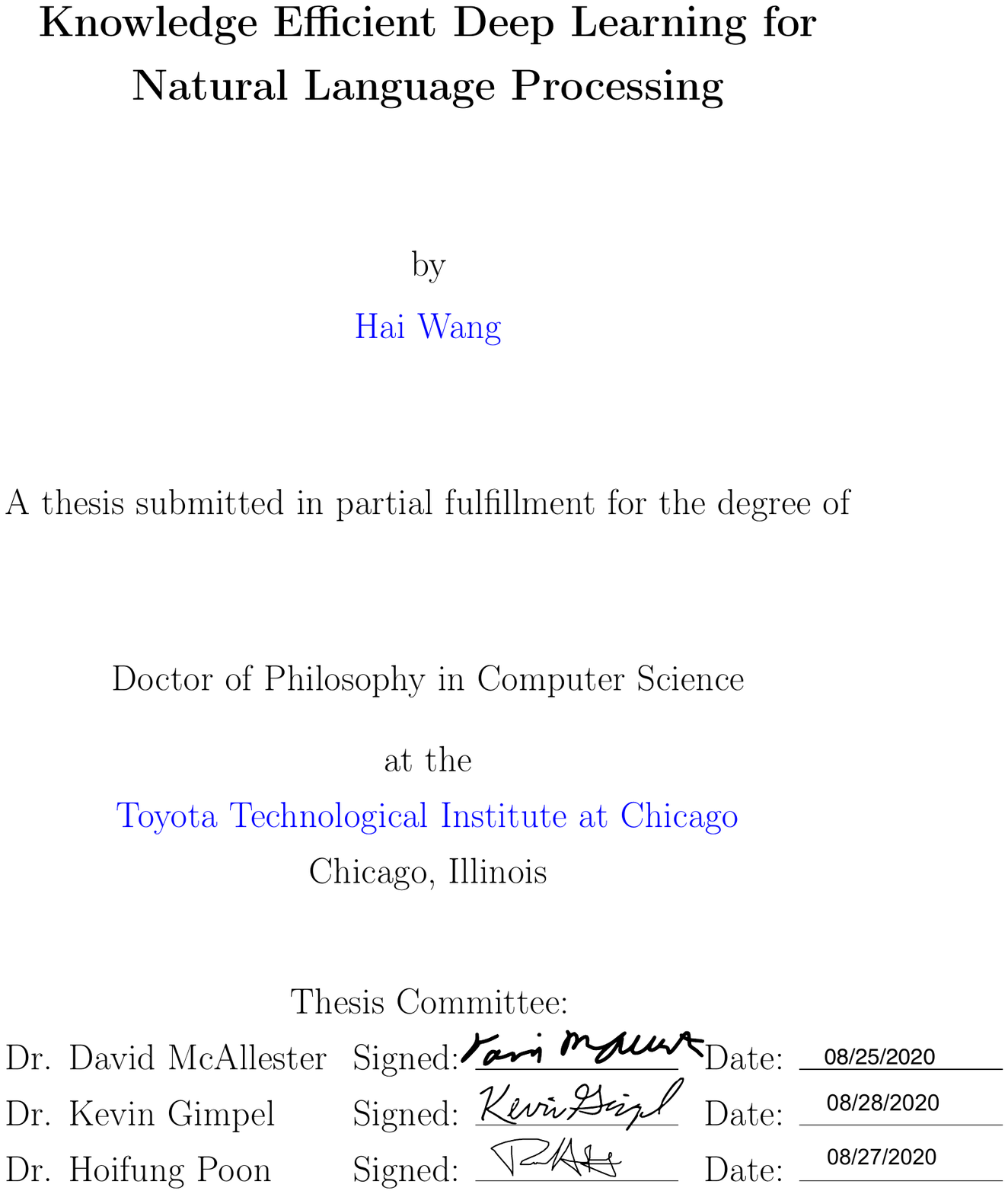}

\ignore{
\Declaration{
\input{Declaration/Declaration.tex}
}
\clearpage  
}

\pagestyle{empty}  

\null\vfill
\textit{``Progress is made by trial and failure; the failures are generally a hundred times more numerous than the successes; yet they are usually left unchronicled.''}

\begin{flushright}
Sir William Ramsay
\end{flushright}

\vfill\vfill\vfill\vfill\vfill\vfill\null
\clearpage  

\addtotoc{Abstract}  
\abstract{
\addtocontents{toc}{\vspace{1em}}  
\input{abstract.tex}
}

\clearpage  

\setstretch{1.3}  

\acknowledgements{
\addtocontents{toc}{\vspace{1em}}  

\input{Acknowledgements/acknowledgements.tex}

}
\clearpage  

\pagestyle{fancy}  

\lhead{\emph{Contents}}  
\tableofcontents  

\lhead{\emph{List of Figures}}  
\listoffigures  

\lhead{\emph{List of Tables}}  
\listoftables  

\setstretch{1.5}  
\clearpage  



\clearpage  
\lhead{\emph{Knowledge Efficient Deep Learning for Natural Language Processing}}  

\ignore{
\listofnomenclature{lll}  
{
$a$ & distance & m \\
$P$ & power & W (Js$^{-1}$) \\
& & \\ 
$\omega$ & angular frequency & rads$^{-1}$ \\
}
}

\setstretch{1.3}  

\pagestyle{empty}  
\dedicatory{This dissertation is dedicated to my parents and my wife. Their support, encouragement, and constant love have sustained me throughout my life.}
\addtocontents{toc}{\vspace{2em}}  

\mainmatter	  
\pagestyle{fancy}  


\newpage
\newpage
\input{introduction.tex} 

\newpage
\input{Chapters/dpl.tex}
\newpage
\input{Chapters/evidence.tex}

\newpage
\input{Chapters/multilingual.tex}
\newpage
\input{Chapters/episodic_memory.tex}
\newpage
\input{conclusion.tex} 


\addtocontents{toc}{\vspace{2em}} 

\appendix 

\input{Appendices/AppendixA}	
\addtocontents{toc}{\vspace{2em}}  
\backmatter

\label{Bibliography}
\lhead{\emph{Bibliography}}  
\bibliographystyle{unsrtnat}  
\bibliography{Bibliography}  

\end{CJK*}
\end{document}

%% file: Declaration/Declaration.tex
\eat
{

\addtocontents{toc}{\vspace{1em}}  

I, Hai Wang, declare that this thesis titled, ``Knowledge Efficient Deep Learning for Natural Language Processing" and the work presented in it are my own. I confirm that:

\begin{itemize} 
\item[\tiny{$\blacksquare$}] This work was done wholly or mainly while in candidature for a Ph.D degree at Toyota Technological Institute at Chicago.

\item[\tiny{$\blacksquare$}] Where any part of this thesis has previously been submitted for a degree or any other qualification at Toyota Technological Institute at Chicago or any other institution, this has been clearly stated.
 
\item[\tiny{$\blacksquare$}] Where I have consulted the published work of others, this is always clearly attributed.
 
\item[\tiny{$\blacksquare$}] Where I have quoted from the work of others, the source is always given. With the exception of such quotations, this thesis is entirely my own work.
 
\item[\tiny{$\blacksquare$}] I have acknowledged all main sources of help.
 
\item[\tiny{$\blacksquare$}] Where the thesis is based on work done by myself jointly with others, I have made clear exactly what was done by others and what I have contributed myself.

\end{itemize}
 
Signed:\\
\rule[1em]{25em}{0.5pt}  
 
Date:\\
\rule[1em]{25em}{0.5pt}  

}

\includepdf[pages=-]{./Declaration/sign_thesis.pdf}

%% file: abstract.tex
Deep learning has become the workhorse for a wide range of natural language processing applications. But much of the success of deep learning relies on annotated examples.  Annotation is time-consuming and expensive to produce at scale. Here we are interested in methods for reducing the required quantity of annotated data --- by making the learning methods more knowledge efficient so as to make them more applicable in low annotation (low resource) settings. There are various classical approaches to making the models more knowledge efficient such as multi-task learning, transfer learning, weakly supervised and unsupervised learning etc. This thesis focuses on adapting such classical methods to modern deep learning models and algorithms.

This thesis describes four works aimed at making machine learning models more knowledge efficient. First, we propose a knowledge rich deep learning model (KRDL) as a unifying learning framework for incorporating prior knowledge into deep models. In particular, we apply KRDL built on Markov logic networks to denoise weak supervision. Second, we apply a KRDL model to assist the machine reading models to find the correct evidence sentences that can support their decision. Third, we investigate the knowledge transfer techniques in multilingual setting, where we proposed a method that can improve pre-trained multilingual BERT based on the bilingual dictionary. Fourth, we present an episodic memory network for language modelling, in which we encode the large external knowledge for the pre-trained GPT.


%% file: Acknowledgements/acknowledgements.tex
First of all, I would like to thank my Ph.D advisor David McAllester without whom I would not be even pursuing a Ph.D in Natural Language Processing at TTIC. David’s foresight for research has reshaped my understanding for research. David also had a great influence in shaping this thesis. Looking back, I think David was a true scientist and the best mentor I could have imaged. I am forever ever grateful for all I have learned from him.

I would like to thank all of my committee members from whom I benefited during my Ph.D journey. I am thankful to Kevin Gimpel who has been helping me when I started doing research in natural language processing and I have benefited a lot from discussions with him. I also want to say thanks to Hoifung Poon for his advices during my internship at Microsoft Research. Choosing this research topic as the final thesis is also partially inspired by him.

I am thankful to Jinbo Xu, my interim advisor who encouraged me to apply to TTIC Ph.D program. Additionally, Madhur Tulsiani deserves a special thanks for being a great director of graduate study program. Madhur’s generous support also made my life at TTIC much easier.

I also want to thank TTIC administrative staffs for helping me whenever I had any questions or encountered problems. First, I would like to thank Chrissy Novak. She always kindly provided suggestions for any problems I had. Adam Bohlander has always been helpful and quick in resolving any GPU server related issues. Amy Minick has always been helpful in answering visa related questions. I also thank Mary Marre, Jessica Johnston and other TTIC staffs for their great effort that makes TTIC is a comfortable place to do research.

I would like to thank many TTIC students and postdocs, I am grateful for knowing them and for all the great moments we shared together. Weiran Wang deserves a special thanks. He was the person I used to unofficially share the office with. I am also thankful for memories I have shared with Zhiyong Wang, Jianzhu Ma, Hao Tang, Haris Angelidakis, Siqi Sun, Heejin Choi, Takeshi Onishi, Mohammadreza Mostajabi, Mrinalkanti Ghosh, Qingming Tang, Lifu Tu, Shubham Toshniwal, Falcon Dai, Zewei Chu and Ruotian Luo. They made my life at TTIC more colorful.

Finally, my deepest gratitude goes to my parents and my wife, they endured an unfair responsibility to support my Ph.D study. Compared to what they have done, any words are barren. Without them, everything in my life would have been totally different.

%% file: introduction.tex
\section{Introduction}

Deep learning has become the main driver of a wide range of NLP tasks ~\cite{bahdanau2014neural,bengio2003neural,clark2016improving,hermann2015teaching,sutskever2014sequence}. 
Deep learning differs from traditional machine learning techniques in that they can automatically learn representations from data such as images, video or text, without the need of introducing hand-coded rules or human domain knowledge. Their highly flexible architectures can allow them to learn directly from raw data and can increase their predictive accuracy when provided with sufficient data~\cite{goodfellow2016deep}. However, success of deep learning is bounded by its reliance on labeled examples, which are expensive and time-consuming to produce. 

To breach the annotation bottleneck and make the deep learning more knowledge efficient, various directions have been proposed, such as multi-task learning~\cite{caruana1997multitask}, transfer learning~\cite{pan2009survey}, few-shot learning~\cite{snell2017prototypical}, unsupervised learning~\cite{le2013building,barlow1989unsupervised,radford2015unsupervised}, weakly and semi-supervised learning~\cite{papandreou2015weakly}. In the context of deep learning, multi-task learning is typically done with either hard or soft parameter sharing of hidden layers between different tasks. Transfer learning assumes that we have pre-trained models used for one task, and we can use those pre-trained models to jump start the training process on a new task. Few-shot learning assumes the model can rapidly generalize from limited supervised experience with few labeled data. Unsupervised learning further assumes no supervision signal to the model at all, instead, the model need to work on its own to discover useful information from the unlabelled data. Semi-supervised learning lies in somewhere between supervised and unsupervised learning. In addition to unlabeled data, semi-supervised learning algorithms are also provided with some supervision information - but not necessarily for all examples. Often, this information will be the labels associated with some of the examples. Weakly supervised learning is more like an umbrella covering several approaches which attempt to build predictive models by learning with various weak supervision.


All those methods have emerged as promising directions to alleviate the annotation bottleneck issue and make the machine learning models more knowledge efficient. Even those methods are not new to the research community, however, the bloom of deep learning creates unique and exciting opportunities for us to revisit them in the context of deep learning. In this thesis, we will review four work we did in this direction.


The first work is a knowledge-rich deep learning model, which is a unified denoising framework for weak supervision~\cite{wang2018deep}. Weak supervision has emerged as a promising direction to address the annotation bottleneck, either by introducing labeling functions to automatically generate noisy examples from unlabeled text, or by imposing constraints over interdependent label decisions. A plethora of methods have been proposed, each with respective strengths and limitations. Probabilistic logic offers a unifying language to represent weak supervision, but end-to-end modeling with probabilistic logic is often infeasible due to intractable inference and learning. In this work, we combine knowledge-rich graphical models with deep learning (KRDL) as a general framework for denosing weak supervision. KRDL models label decisions as latent variables, represents prior knowledge on their relations using weighted first-order logical formulas, and alternates between learning a deep neural network for the end task and refining uncertain formula weights for weak supervision, using variational EM. This framework subsumes prior weak supervision methods as special cases, and enables novel combination via infusion of rich domain and linguistic knowledge. Experiments on biomedical machine reading demonstrate the promise of this approach.

The second work is evidence sentence extraction for machine reading comprehension with the help of knowledge-rich deep learning model~\cite{wang2019evidence}. Since remarkable success has been achieved in the last few years on some machine reading comprehension (MRC) tasks. However, it is still difficult to interpret the predictions of existing MRC models. In this work, we focus on extracting evidence sentences that can explain or support the answers of multiple-choice MRC tasks, where the majority of answer options cannot be directly extracted from reference documents. Due to the lack of ground truth evidence sentence labels in most cases, we apply distant supervision to generate imperfect labels and then use them to train an evidence sentence extractor. To denoise the noisy labels, we apply a recently proposed knowledge rich deep learning framework to incorporate both sentence-level and cross-sentence linguistic indicators for weak supervision. We feed the extracted evidence sentences into existing MRC models and evaluate the end-to-end performance on three challenging multiple-choice MRC datasets: MultiRC, RACE, and DREAM, achieving comparable or better performance than the same models that take as input the full reference document. To the best of our knowledge, this is the first work extracting evidence sentences for multiple-choice MRC.

The third work is investigating knowledge transfer technique in multilingual setting~\cite{wang2019improving}. Recently, pre-trained language models have achieved great success in a broad range of natural language processing tasks. However, in multilingual setting, it is extremely resource-consuming to pre-train a deep language model over large-scale corpora for each language. Instead of exhaustively pre-training monolingual language models independently, an alternative solution is to pre-train a powerful multilingual deep language model over large-scale corpora in hundreds of languages. However, the vocabulary size for each language in such a model is relatively small, especially for low-resource languages. This limitation inevitably hinders the performance of these multilingual models on tasks such as sequence labeling, wherein in-depth token-level or sentence-level understanding is essential. In this work, inspired by previous methods designed for monolingual settings, we investigate two approaches (\ie, joint mapping and mixture mapping) based on a pre-trained multilingual model BERT for addressing the out-of-vocabulary (OOV) problem on a variety of tasks, including part-of-speech tagging, named entity recognition, machine translation quality estimation, and machine reading comprehension. Experimental results show that using mixture mapping is more promising. To the best of our knowledge, this is the first work that attempts to address and discuss the OOV issue in multilingual settings.

The fourth work is large episodic memory language modelling, where we experiment with the use of information retrieval as an augmentation for pre-trained language models. The text corpus used in information retrieval can be viewed as form of episodic memory which grows over time. By augmenting GPT 2.0 with information retrieval we achieve a zero shot 15\% relative reduction in perplexity on Gigaword corpus without any re-training. We also validate our IR augmentation on an event co-reference task.

\eat{
Last, we present our ongoing work, \ie, enriching pre-trained model with vector quantized memory layer.
Recently, large scale pre-trained language models have achieved great success in a broad range of natural language processing tasks. Instead of pre-training BERT like models and using them as encoder for classification tasks, pre-training large scale seq2seq multilingual models for various NLP tasks has become more and more popular. To train such a seq2seq model, the denoise loss (try to recover itself from its corrupted version) is widely used. Naturally, seq2seq model equipped with denoise loss can be formulated as Variational Auto Encoder (VAE) and they will encounter the posterior collapses problem, especially when the decoder in seq2seq model is too powerful. In this work, we borrow the vector quantization idea from Vector Quantizated-VAE (VQ-VAE), and add vector quantization as a special memory layer to large scale pre-trained multilingual seq2seq models. With vector quantization layer, we can circumvent ``posterior collapse" issue, especially for low resource languages. To demonstrate our approach, we first pre-train a small vector quantized multilingual seq2seq model and show its superiority to its original counterpart, then we add vector quantization layer to public available large-scale pre-trained seq2seq models and fine tune them on several downstream tasks. 
}

%% file: Chapters/dpl.tex
\section{Denoising Weak Supervision with Knowledge-Rich Deep Learning}
\label{sec:dpl:paper}

This chapter is based on our previous work ``Deep Probabilistic Logic: A Unifying Framework for Indirect Supervision"~\cite{wang2018deep}. Deep learning has proven successful in a wide range of NLP tasks \cite{bahdanau2014neural,bengio2003neural,clark2016improving,hermann2015teaching,sutskever2014sequence}.
The versatility stems from its capacity of learning a compact representation of complex input patterns \cite{goodfellow2016deep}. 
However, success of deep learning is bounded by its reliance on labeled examples, which are expensive and time-consuming to produce. 
Weak supervision has emerged as a promising direction for breaching the annotation bottleneck.
A powerful paradigm is {\em joint inference} \cite{chang&al07,poon&domingos08,druck&mccallum08,ganchev&al10}, which leverages linguistic and domain knowledge to impose constraints over interdependent label decisions. 
More recently, another powerful paradigm, often loosely called {\em weak supervision}, has gained in popularity. The key idea is to introduce labeling functions to automatically generate (noisy) training examples from unlabeled text.
{\em Distant supervision} is a prominent example that used existing knowledge bases for this purpose \cite{craven1999constructing,mintz2009distant}. {\em Data programming} went further by soliciting labeling functions from domain experts \cite{ratner&al16,bach&al17}. 

Weak-supervision methods have achieved remarkable successes in a number of NLP tasks, but they also exhibit serious limitations.
Distant supervision often produces incorrect labels, whereas labeling functions from data programming vary in quality and coverage, and may contradict with each other on individual instances.
Joint inference incurs greater modeling complexity and often requires specialized learning and inference procedures.

Since these methods draw on diverse and often orthogonal sources of weak supervision, combining them may help address their limitations and amplify their strengths.
Probabilistic logic offers an expressive language for such an integration, and is well suited for resolving noisy and contradictory information \cite{richardson&domingos06}.
Unfortunately, probabilistic logic generally incurs intractable learning and inference, often rendering end-to-end modeling infeasible.

\begin{figure}[!htbp]
    \begin{center}
    \includegraphics[width=0.9\linewidth]{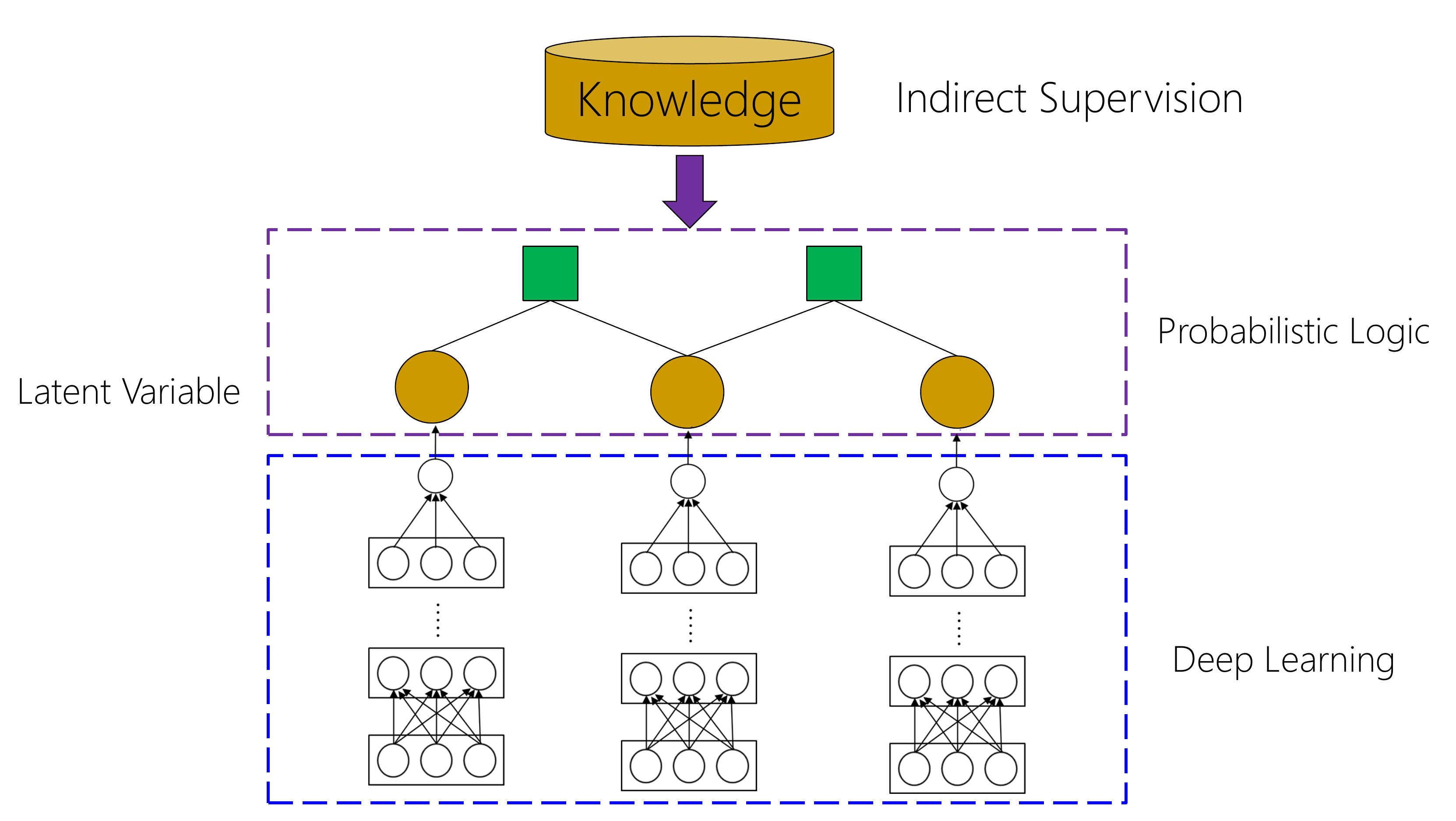}
    \caption{Denoising Weak Supervision with Knowledge-Rich Deep Learning: A general framework for combining weak supervision strategies by composing probabilistic logic with deep learning. Learning amounts to maximizing conditional likelihood of potential function given input by summing up latent label decisions.}
    \label{fig:DPL}
    \end{center}
\end{figure}

In this chapter, we propose {\bf Knowledge-Rich Deep Learning (KRDL)} as a unifying framework for weak supervision (Figure~\ref{fig:DPL}). 
Specifically, we made four contributions.
First, we introduce a modular design to compose probabilistic logic with deep learning, with a supervision module that represents weak supervision using probabilistic logic, and a prediction module that performs the end task using a deep neural network.
Label decisions are modeled as latent variables and serve as the interface between the two modules.

Second, we show that all popular forms of weak supervision can be represented in KRDL by arbitrary potential functions~\cite{subramanya&bilmes07, pearl2014probabilistic}. 
Consequently, these diverse methods can be easily combined within a single framework for mutual amplification.

Third, we show that our problem formulation yields a well-defined learning objective (maximizing conditional likelihood of a potential function).
We proposed a modular learning approach by decomposing the optimization over the supervision and prediction modules, using variational EM, which enables us to apply state-of-the-art methods for probabilistic logic and deep learning.

Finally, we applied KRDL to biomedical machine reading \cite{quirkpoon2017,peng&al17}.
Biomedicine offers a particularly attractive application domain for exploring weak supervision. 

\begin{figure}[!htbp]
    \centering
    \includegraphics[width=0.8\linewidth]{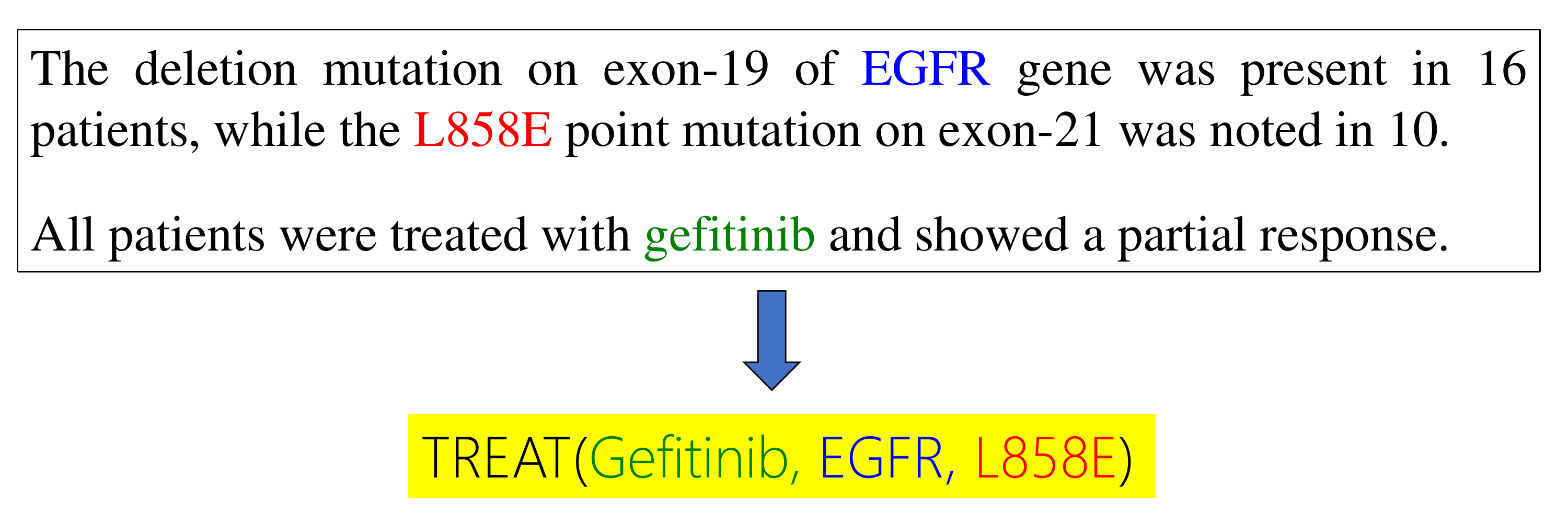}
    \vspace{-5pt}
    \caption{Example of cross-sentence relation extraction for precision cancer treatment.
    }
    \label{fig:relextract-example}
\end{figure}

Biomedical literature grows by over one million articles each year\footnote{\url{http://ncbi.nlm.nih.gov/pubmed}}, making it imperative to develop machine reading methods for automating knowledge curation (Figure~\ref{fig:relextract-example}).
While crowd sourcing is hardly applicable, there are rich domain knowledge and structured resources to exploit for weak supervision.
Using cross-sentence relation extraction and entity linking as case studies, we show that distant supervision, data programming, and joint inference can be seamlessly combined in KRDL to substantially improve machine reading accuracy, without requiring any manually labeled examples.\footnote{The KRDL code and datasets will be made available at \url{ http://hanover.azurewebsites.net}.}

\smallsubsection{Related Work}

\smallpar{Distant supervision}
This paradigm was first introduced for binary relation extraction \cite{craven1999constructing,mintz2009distant}. 
In its simplest form, distant supervision generates a positive example if an entity pair with a known relation co-occurs in a sentence, and samples negative examples from co-occurring entity pairs not known to have the given relation.
It has recently been extended to cross-sentence relation extraction \cite{quirkpoon2017,peng&al17}. In principle, one simply looks beyond single sentences for co-occurring entity pairs. However, this can introduce many false positives and prior work used a small sliding window and filtering (minimal-span) to mitigate training noise. Even so, accuracy is relatively low. 
Both \citet{quirkpoon2017} and \citet{peng&al17} used ontology-based string matching for entity linking, which also incurs many false positives, as biomedical entities are highly ambiguous (e.g., PDF and AAAS are gene names). 
Distant supervision for entity linking is relatively underexplored, and prior work generally focuses on Freebase entities, where links to the corresponding Wikipedia articles are available for learning \cite{huang2015leveraging}.

\smallpar{Data Programming} 
Instead of annotated examples, domain experts are asked to produce labeling functions, each of which assigns a label to an instance if the input satisfies certain conditions, often specified by simple rules \cite{ratner&al16}. 
This paradigm is useful for semantic tasks, as high-precision text-based rules are often easy to come by. 
However, there is no guarantee on broad coverage, and labeling functions are still noisy and may contradict with each other.
The common denoising strategy assumes that labeling functions make random mistakes, and focuses on estimating their accuracy and correlation \cite{ratner&al16,bach&al17}.
A more sophisticated strategy also models instance-level labels and uses instance embedding to estimate instance-level weight for each labeling function \cite{liu2017heterogeneous}.

\smallpar{Joint Inference}
Distant supervision and data programming focus on infusing weak supervision on individual labels. 
Additionally, there is rich linguistic and domain knowledge that does not specify values for individual labels, but imposes hard or soft constraints on their joint distribution.
For example, if two mentions are coreferent, they should agree on entity properties \cite{poon&domingos08}.
There is a rich literature on joint inference for NLP applications. 
Notable methodologies include constraint-driven learning \cite{chang&al07}, general expectation \cite{druck&mccallum08}, posterior regularization \cite{ganchev&al10}, and probabilistic logic \cite{poon&domingos08}. 
Constraints can be imposed on relational instances or on model expectations.
Learning and inference are often tailor-made for each approach, including beam search, primal-dual optimization, weighted satisfiability solvers, etc.
Recently, joint inference has also been used in denoising distant supervision. 
Instead of labeling all co-occurrences of an entity pair with a known relation as positive examples, one only assumes that at least one instance is positive \cite{MultiR,lin&al16}. 

\smallpar{Probabilistic Logic}
Probabilistic logic combines logic's expressive power with graphical model's capability in handling uncertainty. A representative example is Markov logic \cite{richardson&domingos06}, which define a probability distribution using weighted first-order logical formulas as templates for a Markov model.
Probabilistic logic has been applied to incorporating weak supervision for various NLP tasks \cite{poon&domingos07,poon&domingos08,poon&vanderwende10}, but its expressive power comes at a price: learning and inference are generally intractable, and end-to-end modeling often requires heavy approximation 
\cite{kimmig2012short}.
In KRDL, we limit the use of probabilistic logic to modeling weak supervision in the supervision module, leaving end-to-end modeling to deep neural network in the prediction module.
This alleviates the computational challenges in probabilistic logic, while leveraging the strength of deep learning in distilling complex patterns from high-dimension data.

\smallpar{Knowledge-Rich Deep Learning}
Infusing knowledge in neural network training is a long-standing challenge in deep learning \cite{towell1994knowledge}.
\citet{hu2016harnessing,hu2016deep} first used logical rules to help train a convolutional neural network for sentiment analysis. 
KRDL draws inspiration from their approach, but is more general and theoretically well-founded.
\citet{hu2016harnessing,hu2016deep} focused on supervised learning and the logical rules were introduced to augment labeled examples via posterior regularization \cite{ganchev&al10}. 
KRDL can incorporate both direct and weak supervision, including posterior regularization and other forms of weak supervision.
Like KRDL, \citet{hu2016deep} also refined uncertain weights of logical rules, but they did it in a heuristic way by appealing to symmetry with standard posterior regularization. We provide a novel problem formulation using generalized potential function, which shows that their heuristics is a special case of variational EM and opens up opportunities for other optimization strategies.


Deep generative models also combine deep learning with probabilistic models, but focus on uncovering latent factors to support generative modeling and semi-supervised learning \cite{kingma&welling13,kingma&al14}.
Knowledge infusion is limited to introducing structures among the latent variables (e.g., Markov chain) 
\cite{johnson&al16}.
In KRDL, we focus on learning a discriminative model for predicting the latent labels, using a probabilistic model defined by probabilistic logic to inject weak supervision.

\smallsubsection{Denoising Weak Supervision with Knowledge-Rich Deep Learning}
\label{sec:mdl}

In this section, we introduce Knowledge-Rich Deep Learning (KRDL) as an unifying framework for incorporating prior knowledge into deep learning. We will represent prior knowledge as potential functions in a graphical model. The key idea is to model label decisions as latent variables, and introduce a supervision module using a graphical model, which defines a probabilistic distribution over the latent label variables. By combining and denoising weak supervision, we will make the deep learning model more annotation efficient. 

We formulate the learning objective and show how it can be optimized using variational EM, which alternates between estimating marginal probabilities of labels (E-step), as well as using these probabilistic labels to train the deep neural network in the prediction module and refine uncertain parameters in the supervision module (M-step). 


\eat{
\dam{We will consider training data consisting of a set of pairs $(x,y)$ where the task is to predict from $y$ from $x$.
In the relation extraction task described below we will have that $x$ is a sentence with labeled mentions and $y \in \{-1,1\}$ indicates whether a relation holds between the labeled mentions. But in general we can take $x$ and $y$ from arbitrary spaces.}
}

Formally, given a prediction task, let $\mathcal{X}$ denote the set of possible inputs and $\mathcal{Y}$ the set of possible outputs. The goal is to train a prediction module $\Psi(x,y)$ that scores output $y$ given input $x$. In the relation extraction task described later, we will have that $x$ is a sentence with labeled mentions and $y \in \{-1,1\}$ indicates whether a relation holds between the labeled mentions. But in general we can take $x$ from an arbitrary space and take $y$ from a discrete space.

Without loss of generality, we assume that $\Psi(x,y)$ defines the conditional probability $P(y|x)$ using a deep neural network with a softmax layer at the top.
Let $X=(X_1,\cdots,X_N)$ denote a sequence of inputs and $Y=(Y_1,\cdots,Y_N)$ the corresponding outputs. We consider the setting where $Y$ are unobserved, and $\Psi(x,y)$ is learned using weak supervision.

\smallpar{Potential functions} 
We now define $V$ potential functions $(\Phi_1,\cdots,\Phi_V)$ with $\Phi_v(X,Y) \in \mathbb{R}$ defined by  $\Phi_v(X,Y) \propto \exp(w_vf_v(X,Y))$ where $f_v(X,Y) \in \mathbb{R}$ is a feature function represented by a logical formula \cite{richardson&domingos06}.  Here we incorporate expert knowledge by hand designing the feature functions $f_v(X,Y)$.


\smallpar{KRDL}
KRDL comprises of a supervision module over $K = (\Phi_1,\cdots,\Phi_V)$ and a prediction module over all input-output pairs (Figure~\ref{fig:DPL}), and defines a probability distribution:
\vspace{-5pt}
\[P(K,Y|X)\propto \prod_v~\Phi_v(X, Y)\cdot\prod_i~\Psi(X_i, Y_i)\]

\vspace{-8pt}
 A hard constraint is the special case when $w_v=\infty$ (in practice, it suffices to set it to a large number, e.g., 10).
In prior use of potential functions, $w_v$'s are generally pre-determined from prior knowledge. However, this may be suboptimal. Therefore, we consider a general Bayesian learning setting where each $w_v$ is drawn from a pre-specified prior distribution $w_v\sim P(w_v|\alpha_v)$, where $\alpha_v$ is a hyper parameter. Fixed $w_v$ amounts to the special case when the prior is concentrated on the preset value. For uncertain $w_v$'s, we can compute their maximum a posteriori (MAP) estimates and/or quantify the uncertainty.

\smallpar{Distant supervision}
The potential function for distant supervision is similar to that for direct supervision. For example, for relation extraction, distant supervision from a knowledge base of known relations will set $f_{KB}(X_i,Y_i)=\mathbb{I}[\text{\tt In-KB}(X_i,r) \land Y_i=r]$, where $\text{\tt In-KB}(X_i,r)$ is true iff the entity tuple in $X_i$ is known to have relation $r$ in the KB.

\smallpar{Data programming}
Potential functions for data programming are similar to that for distant supervision:
$f_{L}(X_i,Y_i)=\mathbb{I}[L(X_i) = Y_i]$, where $L(X_i)$ is a labeling function provided by domain experts.
Labeling functions are usually high-precision rules, but errors are still common, and different functions may assign conflicting labels to an instance.
Existing denoising strategy assumes that each function makes random errors independently, and resolves the conflicts by weighted votes \cite{ratner&al16}.
In KRDL, this can be done by simply treating error probabilities as uncertain parameters and inferring them during learning. 

\smallpar{Joint inference}
Constraints on instances or model expectations can be imposed by introducing the corresponding potential functions \cite{ganchev&al10} (Proposition 2.1). The weights can be set heuristically \cite{chang&al07,mann&mccallum08,poon&domingos08} or iteratively via primal-dual methods \cite{ganchev&al10}.
In addition to instance-level constraints, KRDL can incorporate arbitrary high-order soft and hard constraints that capture the interdependencies among multiple instances.
For example, identical mentions in proximity probably refer to the same entity, which is useful for resolving ambiguous mentions by leveraging their unambiguous coreferences (e.g., an acronym in apposition of the full name).
This can be represented by the potential functions $f_{\tt Joint}(X_i,Y_i,X_j,Y_j)=\mathbb{I}[{\tt Coref}(X_i,X_j) \land Y_i=Y_j]$, where ${\tt Coref}(X_i,X_j)$ is true iff $X_i$ and $X_j$ are coreferences.
Similarly, the common denoising strategy for distant supervision replaces the mention-level constraints with type-level constraints \cite{MultiR}. 
Suppose that $X_E\subset X$ contains all $X_i$'s with co-occurring entity tuple $E$. The new constraints simply impose that, for each $E$ with known relation $r\in KB$, $Y_i=r$ for at least one $X_i\in X_E$. 
This can be represented by a high-order factor on $(X_i,Y_i: X_i\in X_E)$.

\begin{algorithm}[t]
\begin{algorithmic}
\caption{KRDL Learning}\label{alg:learn}
\State \textbf{Input:} Potential functions $K=\Phi_{1:V}$, deep neural network $\Psi$, inputs $X=(X_1,\cdots,X_N)$.
\State \textbf{Output:} Learned prediction module $\Psi^*$ 
\State \textbf{Initialize:} $\Phi^0 \sim \text{priors}$, $\Psi^0 \sim \text{uniform}$.
\For{\texttt{$t=1:T$}}

      
      \textbf{E step:}
      
      $q^t(Y) = \arg\min_{q}~D_{KL}(\prod_i~q_i(Y_i)~||\prod_v~\Phi^{t-1}_v(X,Y)\cdot\prod_i~\Psi^{t-1}(X_i,Y_i))  $ 
      
      \text{To calculate $q^t(Y)$, we run loopy message pass on graphical model and predict on DL.}
      
      
      \quad \textbf{M step:}
      
       \quad $\Phi^t = 
       \arg\min_{\Phi}~D_{KL}(q^t(Y)~||~ \prod_v~\Phi_v(X,Y)) $
       
       \text{We optimize graphical model $\Phi^t$ with SGD for several epochs using $q^t(Y)$ as soft label. } 
      
      
      
      \quad $ \Psi^t = \arg\min_{\Psi}~D_{KL}(q^t(Y)~||~\prod_i~\Psi(X_i,Y_i)) $ 
      
      \text{We optimize deep learning $\Psi^t$ with SGD for several epochs using $q^t(Y)$ as soft label.}
    
    

\EndFor
\State \Return $\Psi^*=\Psi^T$.
\end{algorithmic}
\end{algorithm}

\smallpar{Parameter learning}
Learning in KRDL maximizes the conditional likelihood of potential functions $P(K|X)$. 
We can directly optimize this objective by summing out latent $Y$ to compute the gradient and run backpropagation.
In this work, however, we opted for a modular approach using variational EM.
See Algorithm~\ref{alg:learn}.

In the E-step, we compute a variational approximation $q(Y)=\prod_i~q_i(Y_i)$ by minimizing its KL divergence with $P(Y|K,X)$, which amounts to computing marginal probabilities $q_i(Y_i)=P(Y_i|K,X)=\sum_{Y_{-i}}~P(Y_i, Y_{-i}|K,X)$, with current parameters $\Phi, \Psi$.
This is a standard probabilistic inference problem. Exact inference is generally intractable, but there are a plethora of approximate inference methods that can efficiently produce an estimate. We use loopy belief propagation \cite{murphy1999loopy} in this work, by conducting message passing in $P(K,Y|X)$ iteratively. Note that this inference problem is considerably simpler than end-to-end inference with probabilistic logic, since the bulk of the computation is encapsulated by 
$\Psi$.

Inference with high-order factors of large size can be challenging, but there is a rich body of literature for handling such structured factors in a principled way. 
In particular, in distant supervision denoising, we alter the message passing schedule so that each at-least-one factor will compute messages to its variables jointly by renormalizing their current marginal probabilities with noisy-or \cite{keith2017identifying}, which is essentially a soft version of dual decomposition \cite{caroe1999dual}.

In the M-step, we treat the variational approximation $q_i(Y_i)$ as probabilistic labels, and use them to optimize $\Phi$ and $\Psi$ via standard supervised learning,  
which is equivalent to minimizing the KL divergence between the probabilistic labels and the conditional likelihood of $Y$ given $X$ under the supervision module ($\Phi$) and prediction module ($\Psi$), respectively.
For the prediction module, this optimization reduces to standard deep learning. 
Likewise, for the supervision module, this optimization reduces to standard parameter learning for log-linear models (i.e., learning all $w_v$'s that are not fixed). Given the probabilistic labels, it is a convex optimization problem with a unique global optimum. Here, we simply use gradient descent, with the partial derivative for $w_v$ being
$\mathbb{E}_{\Phi(Y,X)}~[f_v(X,Y)] - \mathbb{E}_{q(Y)}~[f_v(X,Y)]$.
For a tied weight, the partial derivative will sum over all features that originate from the same template.
The second expectation can be done by simple counting. The first expectation, on the other hand, requires probabilistic inference in the graphical model. But it can be computed using belief propagation, similar to the E-step, except that the messages are limited to factors within the supervision module (i.e., messages from $\Psi$ are not longer included). 
Convergence is usually fast, upon which 
the marginal for each $Y_i$ is available, and $\mathbb{E}_{\Phi(Y,X)}~[f_v(X,Y)]$ is simply the fraction of $Y$ that renders $f_v(X,Y)$ to be true.
Again, this parameter learning problem is much simpler than end-to-end learning with probabilistic logic, as it focuses on refining uncertain weights for weak supervision, rather than learning complex input patterns for label prediction (handled in deep learning).

\begin{figure}[!htbp]
    \centering
    \includegraphics[width=0.95\linewidth]{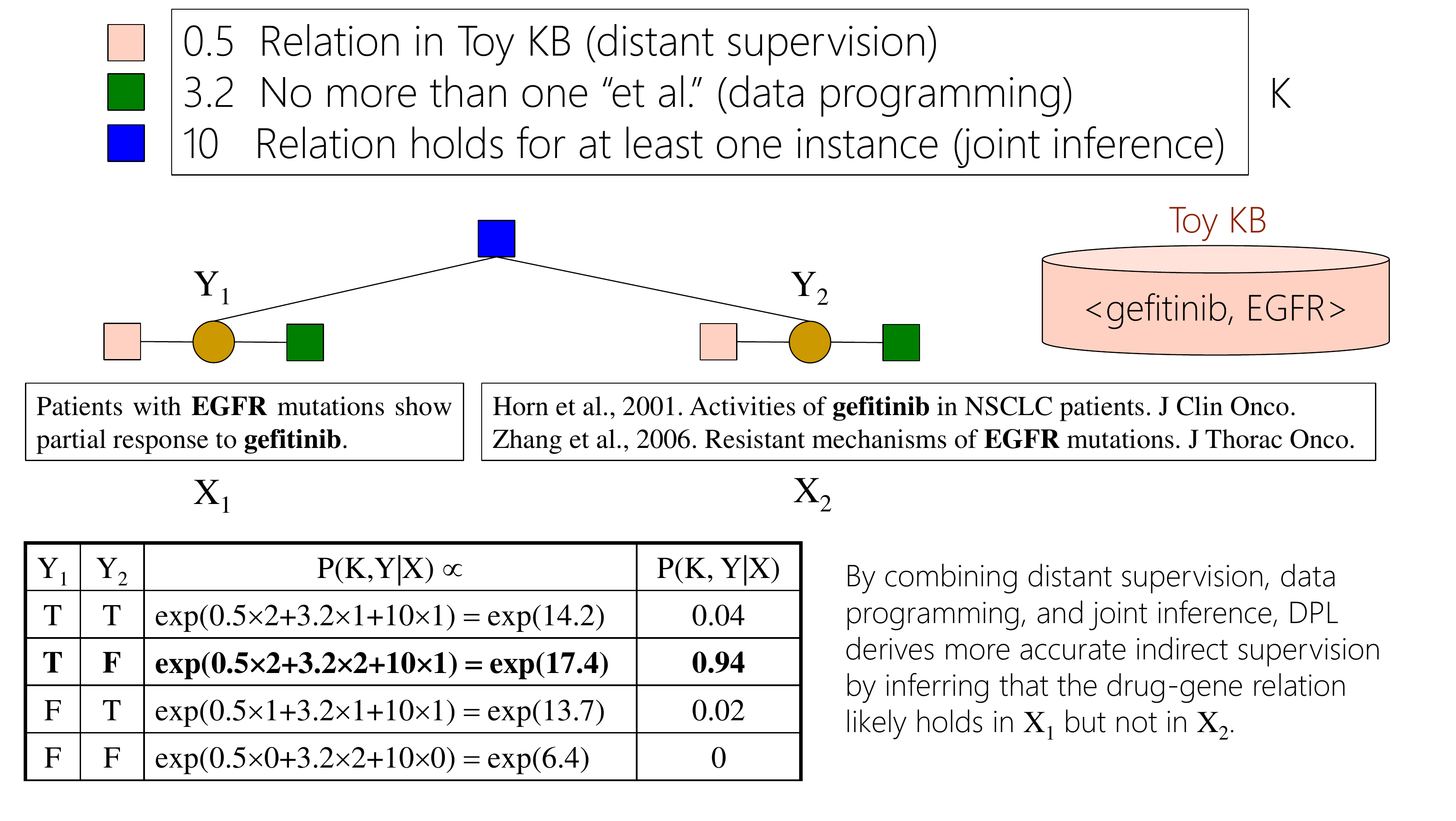}
    \vspace{-10pt}
    \caption{Example of KRDL combining various weak supervision using probabilistic logic. The prediction module is omitted to avoid clutter. 
    }
    \label{fig:dpl-example}
\end{figure}

\smallpar{Example} 
Figure~\ref{fig:dpl-example} shows a toy example on how KRDL combines various weak supervision for predicting drug-gene interaction (e.g., gefitinib can be used to treat tumors with EGFR mutations). Weak supervision is modeled by probabilistic logic, which defines a joint probability distribution over latent labeling decisions for drug-gene mention pairs in unlabeled text. Here, distant supervision prefers classifying mention pairs of known relations, whereas the data programming formula opposes classifying instances resembling citations, and the joint inference formula ensures that at least one mention pair of a known relation is classified as positive. Formula weight signifies the confidence in the weak supervision, and can be refined iteratively along with the prediction module. 

\smallpar{Handling label imbalance} One challenge for distant supervision is that negative examples are often much more numerous. A common strategy is to subsample negative examples to attain a balanced dataset. In preliminary experiments, we found that this was often suboptimal, as many informative negative examples were excluded from training. Instead, we restored the balance by up-weighting positive examples.
In KRDL, an additional challenge is that the labels are probabilistic and change over iterations. In this work, we simply used hard EM, with binary labels set using 0.5 as the probability threshold, and the up-weighting coefficient recalculated after each E-step.

\smallsubsection{Biomedical Machine Reading}

There is a long-standing interest in biomedical machine reading (e.g., \citet{morgan2008overview, kim2009overview}), but prior studies focused on supervised approaches.
The advent of big biomedical data 
creates additional urgency for developing scalable approaches that can generalize to new reading tasks. For example, genome sequencing cost has been dropping faster than Moore's Law, yet oncologists can only evaluate tumor sequences for a tiny fraction of patients, due to the bottleneck in assimilating relevant knowledge from publications.
Recently, \citet{peng&al17} formulated precision oncology machine reading as cross-sentence relation extraction (Figure~\ref{fig:relextract-example}) and developed the state-of-the-art system using distant supervision.
While promising, their results still leave much room to improve.
Moreover, they used heuristics to heavily filter entity candidates, with significant recall loss.

In this section, we use cross-sentence relation extraction as a case study for combining weak supervision using knowledge rich deep learning (KRDL).
First, we show that KRDL can substantially improve machine reading accuracy in a head-to-head comparison with \citet{peng&al17}, using the same entity linking method. 
Next, we apply KRDL to entity linking itself and attain similar improvement.
Finally, we consider further improving the recall by removing the entity filter. 
By applying KRDL to joint entity linking and relation extraction, we more than doubled the recall in relation extraction while attaining comparable precision as \citet{peng&al17} with heavy entity filtering.

\smallpar{Evaluation} 
Comparing weak supervision methods is challenging as there is often
no annotated test set for evaluating precision and recall.
In such cases, we resort to the standard strategy used in prior work by reporting {\em sample precision} (estimated proportion of correct system extractions) and {\em absolute recall} (estimated number of correct system extractions). Absolute recall is proportional to recall and can be used to compare different systems (modulo estimation errors). To guide the learning in preliminary experiments, we use a small annotated set which contains hundreds of examples.

\smallpar{Datasets}
We used the same unlabeled text as \citet{peng&al17}, which consists of about one million full text articles in PubMed Central (PMC)\footnote{\url{www.ncbi.nlm.nih.gov/pmc}}.
Tokenization, part-of-speech tagging, and syntactic parsing were conducted using SPLAT \cite{quirk&al12}, and Stanford dependencies \cite{marneffe&al06} were obtained using Stanford CoreNLP \cite{manning&al14}. 
For entity ontologies, we used DrugBank\footnote{\url{www.drugbank.ca}} and Human Gene Ontology (HUGO)\footnote{\url{www.genenames.org}}. 
DrugBank contains 8257 drugs; we used the subset of 599 cancer drugs.
HUGO contains 37661 genes.
For knowledge bases, we used the Gene Drug Knowledge Database (GDKD)~\cite{dienstmann&al15} and the Clinical Interpretations of Variants In Cancer (CIVIC)\footnote{\url{civic.genome.wustl.edu}}. Together, they contain 231 drug-gene-mutation triples, with 76 drugs, 35 genes and 123 mutations.


\smallsubsection{Cross-sentence relation extraction}

Let $e_1,\cdots,e_m$ be entity mentions in text $T$. 
Relation extraction can be formulated as classifying whether a relation $R$ holds for $e_1,\cdots,e_m$ in $T$.
To enable a head-to-head comparison, we used the same cross-sentence setting as \citet{peng&al17}, where $T$ spans up to three consecutive sentences and $R$ represents the ternary interaction over drugs, genes, and mutations (whether the drug is relevant for treating tumors with the given gene mutation).


\smallpar{Entity linking}
In this subsection, we used the entity linker from Literome \cite{poon&al14} to identify drug, gene, and mutation mentions, as in \citet{peng&al17}. 
This entity linker first identifies candidate mentions by matching entity names or synonyms in domain ontologies, then applies heuristics to filter candidates. The heuristics are designed to enhance precision, at the expense of recall. For example, one heuristics would filter candidates of length less than four, which eliminates key cancer genes such as ER or AKT.

\smallpar{Prediction module}
We used the same graph LSTM as in \citet{peng&al17} to enable head-to-head comparison on weak supervision strategies.
Briefly, a graph LSTM generalizes a linear-chain LSTM by incorporating arbitrary long-ranged dependencies, such as syntactic dependencies, discourse relations, coreference, and connections between roots of adjacent sentences.
A word might have precedents other than the prior word, and its LSTM unit is expanded to include a forget gate for each precedent. 
See \citet{peng&al17} for details.


\begin{table}[t!]
\footnotesize
\begin{center}
\begin{tabular}{ |p{7.5cm} |}
\hline
\textbf{Distant Supervision}: GDKD, CIVIC\\ 
\hline
\textbf{Data Programming (Entity)}\\
\text{Mention matches entity name exactly.}\\
\text{Mention not a stop word.}\\
\text{Mention not following figure designation.}\\
\text{Mention's POS tags indicate it is a noun.}\\
\hline
\textbf{Data Programming (Relation)}\\ 
Less than 30\% of words are numbers in each sentence.\\
No more than three consecutive numbers.\\
No more than two ``et al''.\\
No more than three tokens start with uppercase.\\
No more than three special characters.\\
No more than three keywords indicative of table or figure.\\
Entity mentions do not overlap.\\
\hline 
\textbf{Joint Inference}: \text{Relation holds in at least one instance.} \\ 
\hline
\end{tabular}
\vspace{-5pt}
\caption {KRDL combines three weak supervision strategies for cross-sentence relation extraction}
\label{tb:re_factor}	 
\end{center}
\end{table}

\smallpar{Supervision module}
We used KRDL to combine three weak supervision strategies for cross-sentence relation extraction (Table~\ref{tb:re_factor}).
For distant supervision, we used GDKD and CIVIC as in \citet{peng&al17}.
For data programming, we introduced labeling functions that aim to correct entity and relation errors.
Finally, we incorporated joint inference among all co-occurring instances of an entity tuple with the known relation by imposing the at-least-one constraint (i.e., the relation holds for at least one of the instances).
For development, we sampled 250 positive extractions from KRDL using only distant supervision \cite{peng&al17} and excluded them from future training and evaluation.

\eat{
Maintaining a balanced training set (comparable number in positive and negative examples) is generally a good practice. 
Prior weak supervision methods typically ensure this by sampling a comparable number of negative examples after selecting positive ones. {\bf training instances}
}

\smallpar{Experiment results}
We compared KRDL with the state-of-the-art system of \citet{peng&al17}. 
We also conducted ablation study to evaluate the impact of weak-supervision strategies.
For a fair comparison, we used the same probability threshold in all cases (an instance is classified as positive if the normalized probability score is at least 0.5).
For each system, sample precision was estimated by sampling 100 positive extractions and manually determining the proportion of correct extractions by an author knowledgeable about this domain.
Absolute recall is estimated by multiplying sample precision with the number of positive extractions.

\begin{table}[t]
\centering
\begin{tabular}{ | l | c | c | c |}
\hline
System & Prec. & Abs. Rec.  & Unique \\ \hline
Peng 2017  & 0.64 & 6768 & 2738 \\ \hline
KRDL + $\tt EMB$ & {\bf 0.74} & {\bf 8478} &  {\bf 4821} \\ \hline
KRDL & 0.73 & 7666 &  4144 \\ \hline
~~~$-$ $\tt DS$ & 0.29 & 7555 & 4912  \\ \hline
~~~$-$ $\tt DP$ & 0.67 & 4826 &  2629 \\ \hline
~~~$-$ $\tt DP$ $\tt (ENTITY)$ & 0.70 & 7638 &  4074 \\ \hline
~~~$-$ $\tt JI$ & 0.72 & 7418 & 4011  \\ \hline
\end{tabular}
\vspace{-5pt}
\caption {
Comparison of sample precision and absolute recall (all instances and unique entity tuples) in test extraction on PMC.
KRDL + $\tt EMB$ is our full system using PubMed-trained word embedding, whereas KRDL uses the original Wikipedia-trained word embedding in \citet{peng&al17}. Ablation: DS (distant supervision), DP (data programming), JI (joint inference).
}
\label{tbl:RE-result}
\end{table}

\begin{table}[t]
\begin{center}
\begin{tabular}{ | l | c | c | c |}
\hline
Pred. Mod. & Prec. & \ Abs. Rec. & Unique \\ \hline
BiLSTM  & 0.60 & 6243 &  3427 \\ \hline
Graph LSTM  & 0.73 & 7666 &  4144 \\ \hline
\end{tabular}
\vspace{-5pt}
\caption {Comparison of sample precision and absolute recall (all instances and unique entity tuples) in test extraction on PMC. Both use same weak supervision and Wikipedia-trained word embedding.}
\label{tab:RE_model}
\end{center}
\end{table}

Table~\ref{tbl:RE-result} shows the results. KRDL substantially outperformed \citet{peng&al17}, improving sample precision by ten absolute points and raising absolute recall by 25\%.
Combining disparate weak supervision strategies is key to this performance gain, as evident from the ablation results. While distant supervision remained the most potent source of weak supervision, data programming and joint inference each contributed significantly.
Replacing out-of-domain (Wikipedia) word embedding 
with in-domain (PubMed) word embedding~\cite{pyysalo&al13} also led to a small gain.


\citet{peng&al17} only compared graph LSTM and linear-chain LSTM in automatic evaluation, where distant-supervision labels were treated as ground truth. They found significant but relatively small gains by graph LSTM. We conducted additional manual evaluation comparing the two in KRDL. 
Surprisingly, we found rather large performance difference, with graph LSTM outperforming linear-chain LSTM by 13 absolute points in precision and raising absolute recall by over 20\% (Table~~\ref{tab:RE_model}). 
This suggests that \citet{peng&al17} might have underestimated the performance gain by graph LSTM using automatic evaluation.

\begin{table}[t]
\footnotesize 
\begin{center}
\begin{tabular}{ | p{7.3cm} |}
\hline
\textbf{Distant Supervision}: HGNC\\
\hline 
\textbf{Data Programming}\\
No verbs in POS tags.\\
Mention not a common word. \\
Mention contains more than two characters or one word.\\
More than 30\% of characters are upper case. \\
Mention contains both upper and lower case characters. \\
Mention contains both character and digit. \\
Mention contains more than six characters. \\
Dependency label from mention to parent indicative of direct object.\\

\hline 
\textbf{Joint Inference}\\
Identical mentions nearby probably refer to the same entity.\\
Appositive mentions probably refer to the same entity.\\
Nearby mentions that match synonyms of same entity probably refer to the given entity.
\\ \hline 
\end{tabular}
\vspace{-5pt}
\caption {KRDL combines three weak supervision strategies for entity linking.}
\label{tb:EL-VE}	 
\end{center}
\end{table}
\eat{
-:$\mathbb{I}[{\tt strcase(m)} \land G_m={\tt true}]$; 
-:$\mathbb{I}[{\tt token\_case(m)} \land G_m={\tt true}]$; 
-:$\mathbb{I}[{\tt context\_entropy(m)} \land G_m={\tt true}]$;
-:$\mathbb{I}[{\tt entropy(G_{m})} \land G_m={\tt true}]$; 
-:$\mathbb{I}[{\tt special\_char(m)} \land G_m={\tt true}]$; 
-:$\mathbb{I}[{\tt Acronym(m, text\_span)} \land G_m={\tt true}, \text{apply the heuristics to text\_span}]$  
}


\smallsubsection{Entity linking}

Let $m$ be a mention in text and $e$ be an entity in an ontology. The goal of entity linking is to predict $\tt Link(m,e)$, which is true iff $m$ refers to $e$, for every candidate mention-entity pair $m,e$.
We focus on genes in this work, as they are particularly noisy.


\smallpar{Prediction module} We used BiLSTM with attention over the ten-word windows before and after a mention.
The embedding layer is initialized by word2vec embedding trained on PubMed abstracts and full text~\cite{pyysalo&al13}.
The word embedding dimension was 200. We used 5 epochs for training, with Adam as the optimizer. We set learning rate to 0.001, and batch size to 64. 

\smallpar{Supervision module}
As in relation extraction, we combined three weak supervision strategies using KRDL (Table~\ref{tb:EL-VE}).
For distant supervision, we obtained all mention-gene candidates by matching PMC text against the HUGO lexicon. We then sampled a subset of 200,000 candidate instances as positive examples. We sampled a similar number of noun phrases as negative examples. 
For data programming, we introduced labeling functions that used mention characteristics (longer names are less ambiguous) or syntactic context (genes are more likely to be direct objects and nouns). 
For joint inference, we leverage linguistic phenomena related to coreference (identical, appositive, or synonymous mentions nearby are likely coreferent).
\eat{
\begin{table}[t]
	\begin{center}
		\begin{tabular}{ | l | c | c | c | c| }
			\hline
			System & Acc. & F1 & Prec. & Rec. \\ \hline
			String Match & 0.18 & 0.31 & 0.18 & 1.00 \\ \hline
			DS & 0.64 & 0.43 & 0.28 & 0.90 \\ \hline
			DS + DP   & 0.66 & 0.68 & 0.74 & 0.62\\ \hline
			DS + DP + JI & {\bf 0.73} & {\bf 0.76}   & 0.70 & 0.84   \\ \hline
		\end{tabular}
		\vspace{-5pt}
\caption {Comparison of gene entity linking results on a balanced test set. The string-matching baseline has low precision. By combining weak supervision strategies, KRDL substantially improved precision while retaining reasonably high recall.
}
		\label{tab:EL}	 
	\end{center}
\end{table}
}

\begin{table}[t]
	\begin{center}
		\begin{tabular}{ | l | c | c | c | c| }
			\hline
			System & Acc. & F1 & Prec. & Rec. \\ \hline
			String Match & 0.18 & 0.31 & 0.18 & 1.00 \\ \hline
			DS & 0.64 & 0.71 & 0.62 & 0.83 \\ \hline
			DS + DP   & 0.66 & 0.71 & 0.62 & 0.83\\ \hline
			DS + DP + JI & {\bf 0.70} & {\bf 0.76}   & 0.68 & 0.86   \\ \hline
		\end{tabular}
		\vspace{-5pt}
\caption {Comparison of gene entity linking results on a balanced test set. The string-matching baseline has low precision. By combining weak supervision strategies, KRDL substantially improved precision while retaining reasonably high recall.
}
		\label{tab:EL}	 
	\end{center}
\end{table}


\begin{table}[t]
\begin{center}
\begin{tabular}{ | l | c | c | c |}
\hline
 & F1 & Precision & Recall \\ \hline
GNormPlus  & 0.78 & 0.74 & 0.81 \\ \hline
KRDL  & 0.74 & 0.68 & 0.80 \\ \hline
\end{tabular}
\vspace{-5pt}
\caption {Comparison of gene entity linking results on BioCreative II test set. GNormPlus is the state-of-the-art system trained on thousands of labeled examples. KRDL used only weak supervision.}
\label{tab:GNormPlus}
\end{center}
\end{table}

\smallpar{Experiment results}
For evaluation, we annotated a larger set of sample gene-mention candidates and then subsampled a balanced test set of 550 instances (half are true gene mentions, half not).
These instances were excluded from training and development.
Table~\ref{tab:EL} compares system performance on this test set.
The string-matching baseline has a very low precision, as gene mentions are highly ambiguous, which explains why \citet{peng&al17} resorted to heavy filtering.
By combining weak supervision strategies, KRDL improved precision by over 50 absolute points, while retaining a reasonably high recall (86\%). All weak supervision strategies contributed significantly, as the ablation tests show.
We also evaluated KRDL on BioCreative II, a shared task on gene entity linking  \cite{morgan2008overview}. 
We compared KRDL with GNormPlus \cite{wei2015gnormplus}, the state-of-the-art supervised system trained on thousands of labeled examples in BioCreative II training set.
Despite using zero manually labeled examples, KRDL attained comparable F1 and recall (Table~\ref{tab:GNormPlus}). The difference is mainly in precision, which indicates opportunities for more weak supervision.

\smallsubsection{Joint entity and relation extraction}


An important use case for machine reading is to improve knowledge curation efficiency by offering extraction results as candidates for curators to vet. The key to practical adoption is attaining high recall with reasonable precision~\cite{peng&al17}.
The entity filter used in \citet{peng&al17} is not ideal in this aspect, as it substantially reduced recall. 
In this subsection, we consider replacing the entity filter by the KRDL entity linker Table~\ref{tab:joint}. 
Specifically, we added one labeling function to check if the entity linker returns a normalized probability score above $p_{\tt TRN}$ for gene mentions, and filtered test instances if the gene mention score is lower than $p_{\tt TST}$. 
We set $p_{\tt TRN}=0.6$ and $p_{\tt TST}=0.3$ from preliminary experiments. The labeling function discouraged learning from noisy mentions, and the test-time filter skips an instance if the gene is likely wrong. 
Not surprisingly, without entity filtering, \citet{peng&al17} suffered large precision loss.
All KRDL versions substantially improved accuracy, with significantly more gains using the KRDL entity linker.

\begin{table}[t]
\begin{center}
\begin{tabular}{ | l | c | c | c |}
\hline
System & Prec & Abs. Rec. & Unique \\ \hline
Peng 2017 &	0.31 & 11481 & 5447 \\ \hline
KRDL (RE)  & 0.52 & 17891 & 8534 \\ \hline
~$+$ EL (TRN) & 0.55 & {\bf 21881} &  {\bf 11047} \\ \hline
~$+$ EL (TRN/TST) & {\bf 0.61} & 20378 &  10291 \\ \hline
\end{tabular}
\vspace{-5pt}
\caption{Comparison of sample precision and absolute recall (all instances and unique entity tuples) when all gene mention candidates are considered. \citet{peng&al17} used distant supervision only. RE: KRDL relation extraction. EL: using KRDL entity linking in RE training (TRN) and/or test (TST).
}
\label{tab:joint}	 
\end{center}
\end{table}

\begin{table}[t]
\begin{center}
\begin{tabular}{ | c | c | c| c | c|}
\hline
Gene & Drug & Mut. & Gene-Mut. & Relation\\
\hline
27\% & 4\% & 20\% & 45\% & 24\% \\ \hline
\end{tabular}
\vspace{-5pt}
\caption {Error analysis for KRDL relation extraction.}
\label{tab:error_analyis}	 
\end{center}
\end{table}

\smallsubsection{Discussion}

\smallpar{Scalability} 
KRDL is efficient to train, taking around 3.5 hours for relation extraction and 2.5 hours for entity linking in our PubMed-scale experiments, with 25 CPU cores (for probabilistic logic) and one GPU (for LSTM). For relation extraction, the graphical model of probabilistic logic contains around 7,000 variables and 70,000 factors. At test time, it is just an LSTM, which predicted each instance in less than a second. 
In general, KRDL learning scales linearly in the number of training instances. For distant supervision and data programming, KRDL scales linearly in the number of known facts and labeling functions. As discussed in Section 3, joint inference with high-order factors is more challenging, but can be efficiently approximated. For inference in probabilistic logic, we found that loopy belief propagation worked reasonably well, converging after 2-4 iterations. Overall, we ran variational EM for three iterations, using ten epochs of deep learning in each M-step. We found these worked well in preliminary experiments and used the same setting in all final experiments.

\begin{figure}[!htbp]
    \centering
    \includegraphics[width=0.8\linewidth]{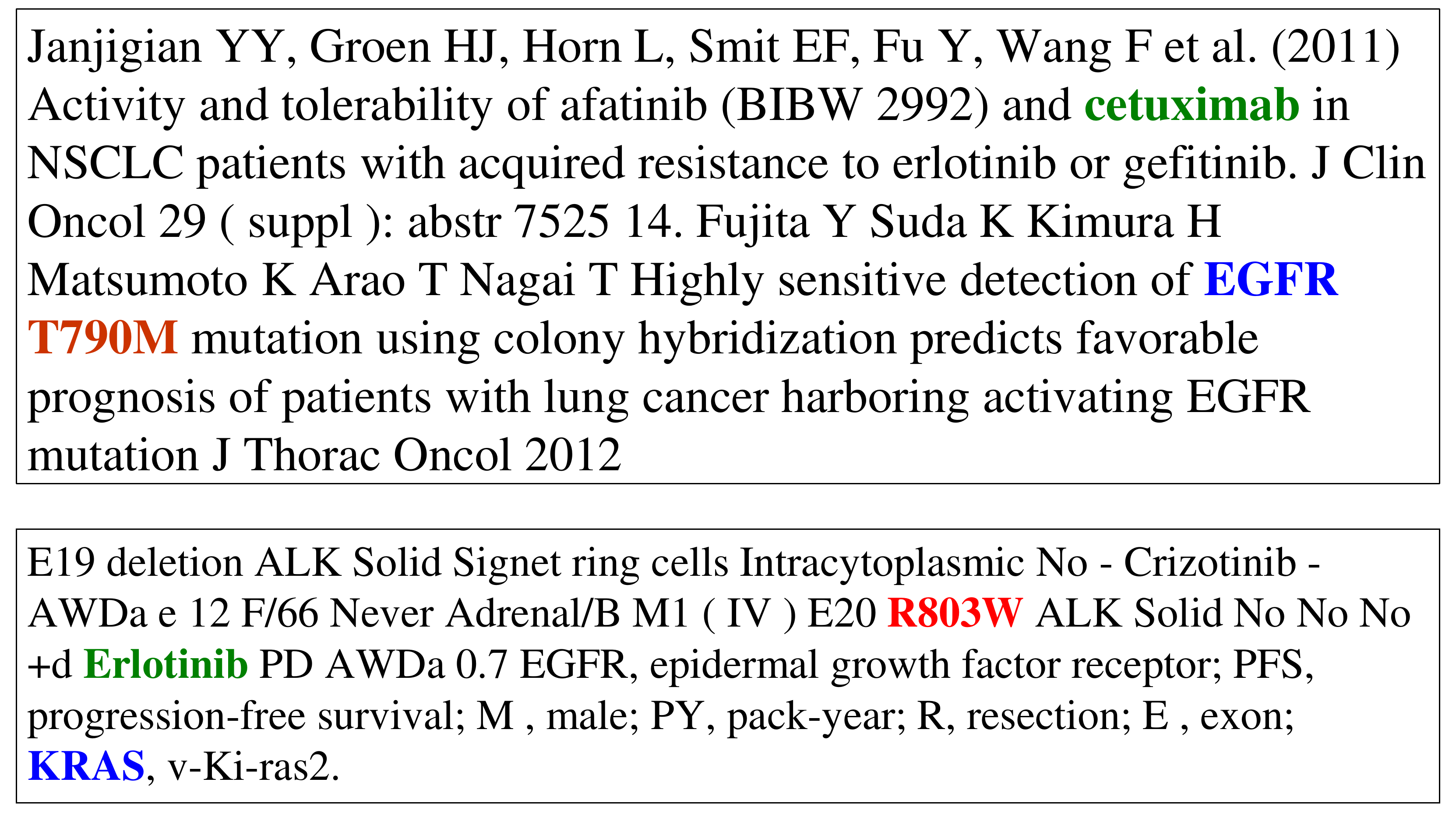}
    \vspace{-5pt}
    \caption{Example of relation-extraction errors corrected by KRDL with additional weak supervision. 
    }
    \label{fig:fix-example}
\end{figure}

\smallpar{Accuracy}
To understand more about KRDL's performance gain over distant supervision, we manually inspected some relation-extraction errors fixed by KRDL after training with additional weak supervision. 
Figure~\ref{fig:fix-example} shows two such examples. 
While some data programming functions were introduced to prevent errors stemming from citations or flattened tables, none were directly applicable to these examples. This shows that KRDL can generalize beyond the original weak supervision.

While the results are promising, there is still much to improve. 
Table~\ref{tab:error_analyis} shows estimated precision errors for relation extraction by KRDL. (Some instances have multiple errors.) 
Entity linking can incorporate more weak supervision.
Joint entity linking and relation extraction can be improved by feeding back extraction results to linking. 
Improvement is also sorely needed in classifying mutations and gene-mutation associations.
The prediction module can also be improved, e.g., by adding attention to graph LSTM.
KRDL offers a flexible framework for exploring all these directions.

\smallsubsection{Conclusion}

We introduce KRDL as a unifying framework for weak supervision, by composing probabilistic logic with deep learning. Experiments on biomedical machine reading show that this enables novel combination of disparate weak supervision methodologies, resulting in substantial gain in accuracy. Future directions include: combining KRDL with deep generative models; exploring alternative optimization strategies; applications to other domains. 

%% file: Chapters/evidence.tex
\section{Evidence Sentence Extraction for Machine Reading Comprehension}

This chapter is based on our previous work ``Evidence Sentence Extraction for Machine Reading Comprehension"~\cite{wang2019evidence}. Recently there have been increased interests in machine reading comprehension (MRC).
We can roughly divide MRC tasks into two categories: 1): extractive/abstractive MRC such as SQuAD~\cite{rajpurkar2016squad}, NarrativeQA~\cite{kovcisky2018narrativeqa}, and CoQA~\cite{reddy2018coqa}; 2): multiple-choice MRC tasks such as MCTest~\cite{richardson2013mctest}, DREAM~\cite{sundream2018} and RACE~\cite{lai2017race}. The MRC tasks in the first category primarily focus on locating text spans from the given reference document/corpus to answer informative factoid questions. In this work, we mainly focus on multiple-choice MRC: given a document and a question, the task aims to select the correct answer option(s) from a small number of answer options associated with this question.   


Existing multiple-choice MRC models~\cite{wang2018co,radfordimproving} take the whole reference document as input and seldom provide evidence snippets, making interpreting their predictions extremely difficult. It is a natural choice for human readers to use several sentences from the reference document to explain why they select a certain answer option in reading tests~\cite{bax2013cognitive}. In this section, as a preliminary attempt, we focus on exacting \emph{\textbf{evidence sentences}} that entail or support a question-answer pair from the reference document and investigating how well a neural reader can answer multiple-choice questions by just using extracted sentences as the input.




From the perspective of evidence sentence extraction, for extractive MRC tasks, information retrieval techniques can already serve as very strong baselines especially when questions provide sufficient information, and most questions are answerable from the content of a single sentence~\cite{lin2018denoising,min2018efficient}. For multiple-choice tasks, there are some unique challenges for evidence sentence extraction. The correct answer options of a significant number of questions (\eg, $87\%$ questions in RACE~\cite{lai2017race,sundream2018}) are not extractive, which require advanced reading skills such as inference over multiple sentences and utilization of prior knowledge~\cite{lai2017race,khashabi2018looking,ostermann2018semeval}. Besides, the existence of misleading distractors (\ie, wrong answer options) also dramatically increases the difficulty of extracting evidence sentences, especially when a question provides insufficient information. For example, in Figure~\ref{fig:overview}, given the reference document and the question \emph{``Which of the following statements is true according to the passage?''}, almost all the tokens in the wrong answer option B \emph{``In 1782, Harvard began to teach German.''} appear in the document (\ie, sentence S$_9$ and S$_{11}$). Furthermore, we notice that even humans sometimes have difficulty in finding pieces of evidence when the relationship between a question and its correct answer option is implicitly indicated in the document (\eg, \emph{``What is the main idea of this passage?''}). Considering these challenges, we argue that extracting evidence sentences for multiple-choice MRC is at least as difficult as that for extractive MRC or factoid question answering.

Given a question, its associated answer options, and a reference document, we propose a method to extract sentences that can support or explain the (question, correct answer option) pair from the reference document. Due to the lack of ground truth evidence sentences in most multiple-choice MRC datasets, inspired by distant supervision, we first select \emph{silver standard} evidence sentences based on the lexical features of a question and its correct answer option (Section~\ref{sec:noisy_gt}), then we use these noisy labels to train an evidence sentence extraction model (Section~\ref{sec:selector}). To denoise the distant supervision, we leverage rich linguistic knowledge from external resources such as ConceptNet~\cite{speer2017conceptnet} and Paraphrase Database~\cite{pavlick2015ppdb}, and we accommodate all those weak supervision with a recently proposed knowledge rich deep learning~\cite{haidpl2018} framework (Section~\ref{sec:dpl}). We combine our evidence extractor with two recent neural readers~\cite{wang2018co,radfordimproving} and evaluate the end-to-end performance on three challenging multiple-choice MRC datasets: MultiRC~\cite{khashabi2018looking}, DREAM~\cite{sundream2018}, and RACE~\cite{lai2017race}. Experimental results show that we achieve comparable or better performance than baselines that consider the full context, indirectly demonstrating the quality of our extracted sentences. We also compare our evidence extractor with a recently proposed sentence selector~\cite{lin2018denoising}. Our extractor significantly outperforms the baseline selector in filtering out noisy retrieved paragraphs on two open-domain factoid question answering datasets: Quasar-T~\cite{dhingra2017quasar} and SearchQA~\cite{dunn2017searchqa}. 


Our primary contributions are as follows: 1) to the best of our knowledge, we present the first work to extract evidence sentences for multiple-choice MRC; 2) we utilize various sources of weak supervision derived from linguistic knowledge to denoise the noisy evidence sentence labels and demonstrate the value of linguistic knowledge for MRC. We hope our attempts and observations can encourage the research community to develop more explainable models that simultaneously provide predictions and textual evidence. 

\subsection{Related Work}

\textbf{Sentence Selection for MRC/Fact Verification}
Previous studies investigate paragraph retrieval for factoid question answering~\cite{chen2017reading,wang2017r,choi2017coarse,lin2018denoising}, sentence selection for machine reading comprehension~\cite{hewlett2017accurate,min2018efficient}, and fact verification~\cite{yin2018twowingos,hanselowski2018ukp}. In these tasks, most of the factual questions/claims provide sufficient clues for identifying relevant sentences, thus often information retrieval combined with filters can serve as a very strong baseline. For example, in the FEVER dataset~\cite{thorne2018fever}, only $16.8\%$ of claims require composition of multiple evidence sentences. Different from above work, we exploit information in answer options and use various weak supervision to train our evidence extractor, and previous work can actually be a regarded as a special case for our pipeline. Compared to~\citet{lin2018denoising}, we leverage rich linguistic knowledge for denoising. 

Several work also investigate content selection at the token level~\cite{yu2017learning,seo2017neural}, in which some tokens are automatically skipped by neural models. However, they do not utilize any linguistic knowledge, and a set of discontinuous tokens has limited explanation capability.

\textbf{MRC with External Knowledge} Linguistic knowledge such as coreference resolution, frame semantics, and discourse relations is widely used to improve machine comprehension~\cite{wang2015machine,sachan2015learning,narasimhan2015machine,sun2018reading} especially when there are only hundreds of documents available in a dataset such as MCTest~\cite{richardson2013mctest}. 
Along with the creation of large-scale reading comprehension datasets, recent MRC models rely on end-to-end neural models, and it primarily uses word embeddings as input. However, ~\citet{wang2016emergent,dhingra2017linguistic, dhingra2018neural} show that existing neural models do not fully take advantage of the linguistic knowledge, which is still valuable for MRC. 
Besides widely used lexical features such as part-of-speech tags and named entity types~\cite{wang2016emergent,liu2017stochastic,dhingra2017linguistic, dhingra2018neural}, we consider more diverse types of external knowledge for performance improvements. Moreover, we accommodate external knowledge with probabilistic logic to potentially improve the interpretability of MRC models instead of using external knowledge as additional features.

\textbf{Explainable MRC/Question Answering} To improve the interpretability of question answering, previous work utilize interpretable internal representations~\cite{palangi2017question} or reasoning networks that employ a hop-by-hop reasoning process dynamically~\cite{zhou2018interpretable}. A research line focuses on visualizing the whole derivation process from the natural language utterance to the final answer for question answering over knowledge bases~\cite{abujabal2017quint} or scientific word algebra problems~\cite{ling2017program}. ~\citet{jansen2016s} extract explanations that describe the inference needed for elementary science questions (\eg, \emph{``What form of energy causes an ice cube to melt''}). In comparison, the derivation sequence is less apparent for open-domain questions, especially when they require external domain knowledge or multiple-sentence reasoning. To improve explainability, we can also check the attention map learned by neural readers~\cite{wang2016emergent}, however, attention map is learned in end-to-end fashion, which is different from our work.

A similar work proposed by~\cite{sharp2017tell} also uses distant supervision to learn how to extract informative justifications. However, their experiments are primarily designed for factoid question answering, in which it is relatively easy to extract justifications since most questions are informative. In comparison, we focus on multi-choice machine reading comprehension that requires deep understanding, and we pay particular attention to denoising strategies.

\subsection{Method}

\begin{figure}[!htbp]
   \begin{center}
   \includegraphics[width=0.98\textwidth]{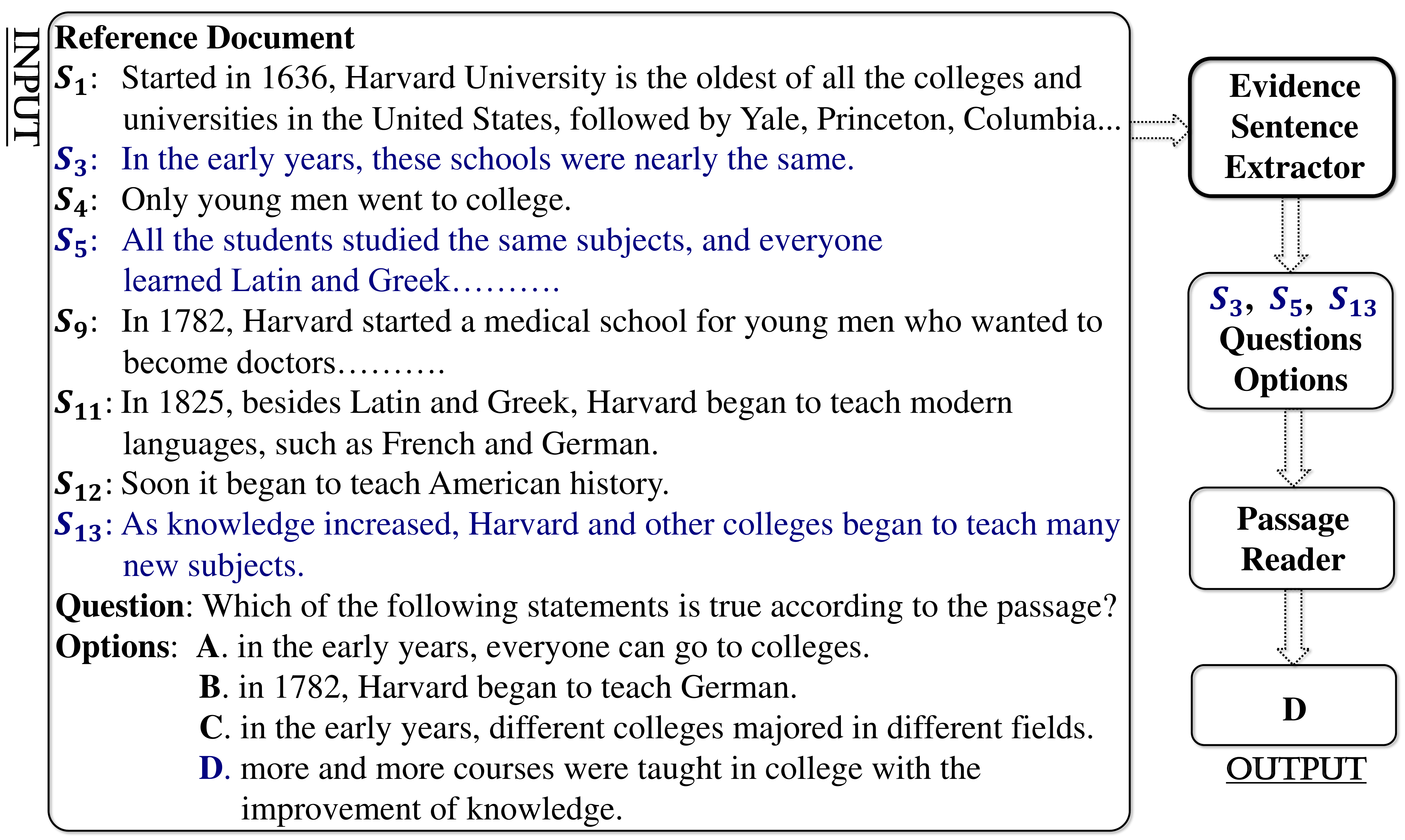}
   \end{center}
 \caption{An overview of our pipeline. The input instance comes from RACE~\cite{lai2017race}.}
 \label{fig:overview}
\end{figure}

Our pipeline contains a \emph{\textbf{neural evidence extractor}} trained on the noisy training data generated by distant supervision and an existing \emph{\textbf{neural reader}} for answer prediction that takes evidence sentences as input. We detail the entire pipeline in Section~\ref{sec:methods} and show an overview in Figure~\ref{fig:overview}.

\subsubsection{Transformer}
\label{sec:transformer}

We primarily use a multi-layer multi-head transformer~\cite{vaswani2017attention} to extract evidence sentences.
Let $W_{w}$ and $W_{p}$ be the word (subword) and position embeddings, respectively. Let $M$ denote the total number of layers in the transformer. Then, the $m$-th layer hidden state $h^{m}$ of a token is given by:

\begin{equation}
\small
h^m = \begin{cases}
W_{w} + W_{p} &\text{if $m=0$}\\
\text{TB}(h^{m-1})  &\text{if $1\leq m \leq M$}
\end{cases}
\end{equation}
where TB stands for the transformer block, which is a standard module that contains MLP, residual connections~\cite{he2016deep}, self attention~\cite{vaswani2017attention} and LayerNorm~\cite{ba2016layer}.

Recently, several pre-trained transformers such as GPT~\cite{radfordimproving}
and BERT~\cite{devlin2018bert} have been released. Compared to RNNs such as LSTMs~\cite{hochreiter1997long} and GRUs~\cite{cho2014learning}, pre-trained transformers capture rich world and linguistic knowledge from large-scale external corpora, and significant improvements are obtained by fine-tuning these pre-trained models on several downstream tasks. We follow this promising direction by fine-tuning GPT~\cite{radfordimproving}. Note that the pre-trained transformer in our pipeline can also be easily replaced by BERT.

We use $(X, Y)$ to denote all training data, $(X_{i}, Y_{i})$ to denote each instance, where $X_{i}$ is a token sequence, namely, $X_{i} = \{X_{i}^{1},\ldots, X_{i}^{t} \}$ where $t$ equals to the sequences length. For evidence extraction, $X_{i}$ contains one sentence in a document, a question, and all answer options associated with the question. $Y_{i}$ indicates the probability that sentence $X_{i}$ is selected as an evidence sentence for this question, and $\sum_{i=1}^{N}Y_{i}=1$ where $N$ equals to the total number of sentences in a document. The transformer takes $X_{i}$ as input and produces the final hidden state $h_{i}^{M}$ of the last token in $X_{i}$~\cite{radfordimproving}, which is further fed into a linear layer followed by a softmax layer to generate the probability:
\begin{eqnarray}
 P_{i} = \frac{\text{exp}(W_{y}h_{i}^{M})}{\sum_{1 \leq i \leq N}\text{exp}(W_{y}h_{i}^{M})}
\end{eqnarray}
where $W_{y}$ is the vector for the output layer. Kullback-Leibler (KL) divergence loss $\text{KL}((Y_{1}, \ldots, Y_{N})||(P_{1}, \ldots, P_{N}))$ is used as training criteria.

\eat{
KL divergence is used as the training criteria:
\begin{eqnarray}
 L = \sum_{x, y} \text{KL} \Big \{ p(y \, | \, x_{1},\ldots, x_{t}), p(y) \Big\}
\end{eqnarray}
}

\subsubsection{Knowledge Rich Deep Learning}
\label{sec:dpl}

Since human-labeled evidence sentences are seldom available in existing machine reading comprehension datasets, we use distant supervision to generate weakly labeled evidence sentences: we know the correct answer options, then we can select the sentences in the reference document that have the highest information overlapping with the question and the correct answer option. However, weakly labeled data generated by distant supervision is inevitably noisy~\cite{bing2015improving}, and therefore we need a denoising strategy that can leverage various sources of weak supervision. 

In this section, we use Knowledge Rich Deep Learning (KRDL) which introduced in Section \ref{sec:dpl:paper}, which is an unifying denoise framework that can efficiently model various weak supervision by integrating probabilistic logic with deep learning. To recap, it consists of two modules: 1) a supervision module that represents weak supervision using probabilistic logic; 2) a prediction module that uses deep neural networks to perform the downstream task. The label decisions derived from weak supervision are modeled as latent variables and serve as the interface between the two modules. KRDL combines three sources of weak supervision: data programming, distant supervision, and joint inference. For data programming, we introduce a set of labeling functions that are specified by simple rules and written by domain experts, and each function assigns a label to an instance if the input satisfies certain conditions. We will detail these sources of weak supervision under our task setting in Section~\ref{sec:methods}.

\subsubsection{Our Pipeline}
\label{sec:methods}

\begin{figure*}[!htbp]
   \begin{center}
   \includegraphics[width=1.0\textwidth]{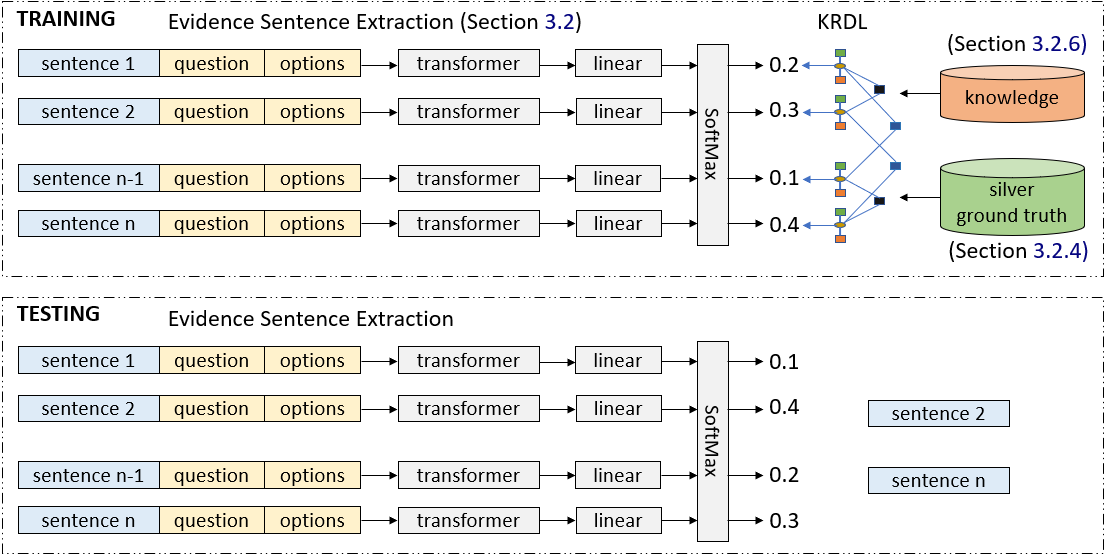}
   \end{center}
 \caption{Knowledge rich deep learning framework for evidence extraction. At test time, we only use trained neural evidence extractor for prediction.}
 \label{fig:method:nnStruct}
\end{figure*}

As shown in Figure \ref{fig:method:nnStruct}, in training stage, our evidence extractor contains two components: a probabilistic graph containing various sources of weak supervision used as a supervision module (Section~\ref{sec:dpl}) and a fine-tuned pre-trained transformer used as a prediction module. The two components are connected via a set of latent variables indicating whether each sentence is an evidence sentence or not. We update the model by alternatively optimizing the transformer and the probabilistic graph so that they reach an agreement on latent variables. After training, only the transformer is kept to make predictions for a new instance during testing.

As we mentioned in Section~\ref{sec:dpl}, KRDL can jointly represent three sources of different weak supervision. We first introduce two distant supervision methods to generate noisy evidence sentence labels (Section~\ref{sec:noisy_gt}). We then introduce other sources of weak supervision --- data programming and joint inference --- used for denoising in KRDL (Section~\ref{sec:selector}).

\subsubsection{Silver Standard Evidence Generation}
\label{sec:noisy_gt}

Given correct answer options, we use two different distant supervision methods to generate the \emph{silver standard} evidence sentences.  

\noindent \textbf{Rule-Based Method} We select sentences that have higher weighted token overlap with a given (question, correct answer options) pair as silver standard evidence sentences. Tokens are weighted by the inverse term frequency.

\noindent \textbf{Integer Linear Programming (ILP)}
Inspired by ILP models for summarization~\cite{berg2011jointly,boudin2015concept}, we model evidence sentence selection as a maximum coverage problem and define the value of a selected sentence set as the sum of the weights for the unique words it contains. Formally, let $v_i$ denote the weight of word $i$, $v_i=1$ if word $i$ appears in the correct answer option,  $v_i=0.1$ if it appears in the question but not in the correct answer option, and $v_i=0$ otherwise.\footnote{We do not observe a significant improvement by tuning parameters $v_i$ on the development set.}

We use binary variables $c_i$ and $s_j$ to indicate the presence of word $i$ and sentence $j$ in the selected sentence set, respectively. $\text{Occ}_{i,j}$ is a binary variable indicating the occurrence of word $i$ in sentence $j$, $l_j$ denotes the length of sentence $j$, and $L$ is the predefined maximum number of selected sentences. We formulate the ILP problem as: 
\begin{eqnarray}
\small
 \max \sum_i v_i c_i \quad  s.t. \sum_j s_j \leq L \\
 s_j ~\text{Occ}_{ij} \leq c_i,  \forall i,j \quad \sum_{j} s_j ~\text{Occ}_{ij} \geq c_i, \forall i \\ \nonumber
c_i \in \{0, 1\} \ \forall i \quad s_j \in \{0, 1\} \ \forall j \nonumber
\end{eqnarray}

\subsubsection{Denoising with knowledge rich deep learning}
\label{sec:selector}

Besides distant supervision, KRDL also includes data programming and joint inference (\ie, $f_{v}(X, Y)$ in Section~\ref{sec:dpl}). As a preliminary attempt, we manually design a small number of sentence-level labeling functions for data programming and high-order factors for joint inference. We briefly introduce them as follows and list the implementation details in Section~\ref{sec:factor}.

For sentence-level functions, we consider lexical features (\ie, the sentence length, the entity types in a sentence, and sentence positions in a document), semantic features based on word and paraphrase embeddings and ConceptNet~\cite{speer2017conceptnet} triples, and rewards for each sentence from an existing neural reader, language inference model, and sentiment classifier, respectively.

For high-order factors, we consider factors including if whether adjacent sentences prefer the same label, the maximum distance between two evidence sentences that support the same question, and the token overlap between two evidence sentences that support different questions.

\begin{figure}[!htbp]
   \begin{center}
   \includegraphics[width=0.75\textwidth]{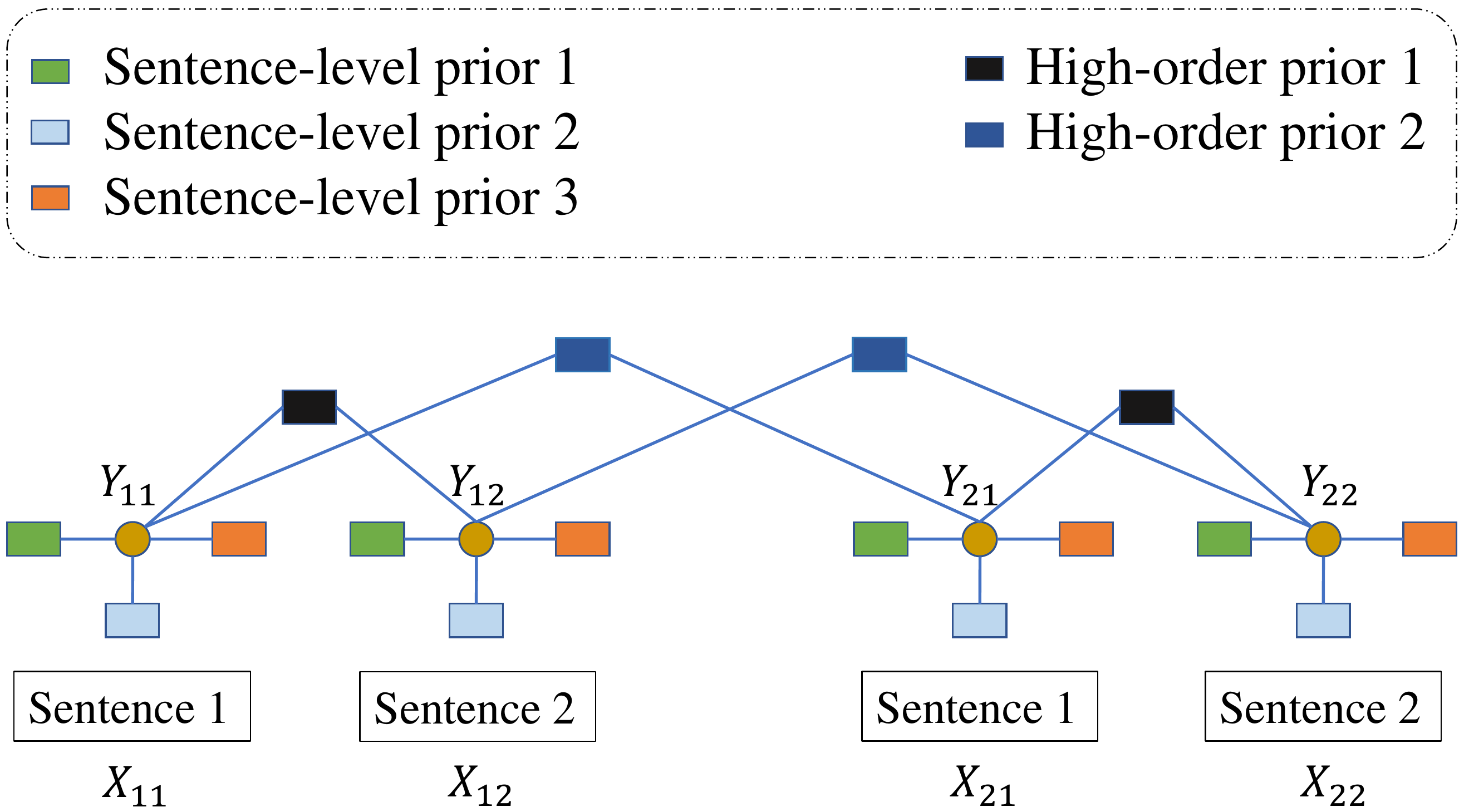}
   \end{center}
\caption{A simple factor graph for denoising.}
\label{fig:factorgraph}
\end{figure}

We show the factor graph for a toy example in Figure~\ref{fig:factorgraph}, where the document contains two sentences and two questions. $X_{ij}$ denotes an instance consisting of sentence $i$, question $j$ and its associated options, $Y_{ij}$ is a latent variable indicating the probability that sentence $i$ is an evidence sentence for question $j$. We build a factor graph for the document and all its associated questions jointly. By introducing the logic rules jointly over $X_{ij}$ and $Y_{ij}$, we can model the joint probability for $Y$.

\subsubsection{Factors for Denoising}
\label{sec:factor}

Besides distant supervision, KRDL also includes data programming and joint inference. For data programming, we design the following sentence-level labeling functions:

\textbf{Sentence-Level Labeling Functions}
\label{app:low}

\begin{itemize}
\setlength\itemsep{-0.25em}
\item Sentences contain the information asked in a question or not: for ``when"-questions, a sentence must contain at least one time expression; for ``who"-questions, a sentence must contain at least one person entity.  
\item Whether a sentence and the correct answer option have a similar length: $0.5 \leq \frac{\text{len(sentence)}}{\text{len(answer)}} \leq 3$.
\item A sentence that is neither too short nor too long since those sentences tend to be less informative or contain irrelevant information: $5 \leq \#~\text{of tokens in sentence} \leq 40$.
\item Reward for each sentence from a neural reader. We sample different sentences and use their probabilities of leading to the correct answer option as rewards. See Section~\ref{sec:implementation} for details about reward calculation.
\item Paraphrase embedding similarity between a question and each sentence in a document: $ \cos({e^{para}_{q}, e^{para}_{sent}}) \geq 0.4 $. 
\item Word embedding similarity between a question and each sentence in a document: $ \cos({e^{w}_{q}, e^{w}_{sent}}) \geq 0.3 $.
\item Whether question and sentence contain words that have the same entity type.
\item Whether a sentence and the question have the same sentiment classification result.
\item Language inference result between sentence and question: entail, contradiction, neutral.
\item \# of matched tokens between the concatenated question and candidate sentence with the triples in ConceptNet~\cite{speer2017conceptnet}: $\frac{\#~\text{of matching}}{\#~\text{of tokens in sentence}} \leq 0.2$.  
\item If a question requires the document-level understanding, we prefer the first or the last three sentences in the reference document.
\end{itemize}

\textbf{High-Order Factors}
\label{app:high}

For joint inference, we consider the following high-order factors $f_{v}(X, Y)$.
\begin{itemize}
\setlength\itemsep{-0.25em}
    \item Adjacent sentences prefer the same label.
    \item Evidence sentences for the same question should be within window size $8$. For example, we assume $S_1$ and $S_{12}$ in Figure~\ref{fig:overview} are less likely to serve as evidence sentences for the same question.
    \item Overlap ratio between evidence sentences for different questions is smaller than $0.5$. We assume the same set of evidence sentences are less likely to support multiple questions.
\end{itemize}

\subsection{Datasets}
\label{sec:dataset}


\begin{table}[ht!]
\footnotesize
\centering
\begin{tabular}{lccccccc}
\toprule
\multirow{2}{*}{\textbf{Dataset}} & \multicolumn{3}{c}{\textbf{\# of documents}} & \multicolumn{3}{c}{\textbf{\# of questions}} & \textbf{Average \# of sentences per document} \\
                         & Train& Dev & Test& Train & Dev & Test & Train + Dev + Test \\
\midrule
MultiRC                  &  456       & 83         & 332     & 5,131       &  953  & 3,788 & 14.5 (Train + Dev)    \\
DREAM                    & 3,869      & 1,288     & 1,287     & 6,116       & 2,040        & 2,041 & -\\
RACE                     & 25,137     & 1,389     & 1,407     & 87,866      & 4,887        & 4,934  & 17.6   \\
\midrule
Quasar-T                     & -    & -   & -     & 37,012      & 3,000        & 3,000  & 100   \\
SearchQA                     & -    & -   & -   & 99,811       & 13,893        & 27,247 & 50    \\
\bottomrule
\end{tabular}
\caption{Statistics of multiple-choice machine reading comprehension and question answering datasets.}
\label{tab:dataset_statistics}
\end{table}

We primarily focus on extracting evidence sentences for multiple-choice machine reading comprehension. Three latest MRC datasets are investigated (Section~\ref{sec:multi}). Additionally, to have a head-to-head comparison with existing sentence selectors designed for factoid question answering, we also evaluate our approach on two open-domain question answering datasets, in which answers are text spans (Section~\ref{sec:qa}). See Table~\ref{tab:dataset_statistics} for statistics.

\subsubsection{Multiple-Choice Datasets}
\label{sec:multi}

\textbf{MultiRC}~\cite{khashabi2018looking}: MultiRC is a dataset in which questions can only be answered by considering information from multiple sentences. There can exist multiple correct answer options for a question. Reference documents come from seven different domains such as elementary school science and travel guides. For each document, questions and their associated answer options are generated and verified by turkers.


\noindent \textbf{DREAM}~\cite{sundream2018}: DREAM is a dataset collected from English Listening exams for Chinese language learners. Each instance in DREAM contains a multi-turn multi-party dialogue, and the correct answer option must be inferred from the dialogue context. In particular, a large portion of questions require multi-sentence inference ($84\%$) and/or commonsense knowledge ($34\%$). 

\noindent \textbf{RACE}~\cite{lai2017race}: RACE is a dataset collected from English reading exams designed for middle (RACE-Middle) and high school (RACE-High) students in China, carefully designed by English instructors. The proportion of questions that requires reasoning is $59.2\%$.


\subsubsection{Question Answering Datasets}
\label{sec:qa}

\textbf{Quasar-T}~\cite{dhingra2017quasar}: It contains open-domain questions and their associated answers extracted from ClueWeb09. For each question, $100$ sentences are retrieved from ClueWeb09 using information retrieval techniques.

\noindent \textbf{SearchQA}~\cite{dunn2017searchqa}: For each question,~\citet{dunn2017searchqa} retrieve $50$ web pages from J! Archive as the relevant documents using the Google Search API.

\subsubsection{Implementation Details}
\label{sec:implementation}

We use spaCy~\cite{honnibal2015improved} for tokenization and named entity tagging. We use the pre-trained transformer released by~\citet{radfordimproving} with the same pre-processing procedure. When the transformer is used as the neural reader, we set training epochs to 4, use eight P40 GPUs for experiments on RACE, and use one GPU for experiments on other datasets. When the transformer is used as the evidence extractor, we set batch size 1 per GPU and dropout rate $0.3$. We keep other parameters default. Depending on the dataset, training the evidence extractor generally takes several hours. Training neural readers with evidence sentences as input takes significant less time than that with full context as input.

For KRDL, we adopt the toolkit from~\citet{haidpl2018}. We use Vader~\cite{gilbert2014vader} for sentiment analysis and ParaNMT-$50$M~\cite{wieting2018paranmt} to calculate the paraphrase similarity between two sentences. We use the triples in ConceptNet v$5.6$~\cite{speer2012representing,speer2017conceptnet} to incorporate commonsense knowledge. To calculate the natural language inference probability, we first fine-tune the transformer~\cite{radfordimproving} on several tasks, including SNLI~\cite{bowman2015large}, SciTail~\cite{khot2018scitail}, MultiNLI~\cite{williams2018broad}, and QNLI~\cite{wang2018glue}.

To calculate the probability that each sentence leads to the correct answer option, we sample a subset of sentences and use them to replace the full context in each instance, and then we feed them into the transformer fine-tuned with instances with full context. If a particular combination of sentences $S=\{s_{1},\ldots,s_{n}\}$ leads to the prediction of the correct answer option, we reward each sentence inside this set with $1/n$. To avoid the combinatorial explosion, we assume evidence sentences lie within window size $3$. For another neural reader Co-Matching~\cite{wang2018co}, we use its default parameters. For DREAM and RACE, we set $L$, the maximum number of silver standard evidence sentences of a question, to $3$. For MultiRC, we set $L$ to 5 since many questions have more than $5$ ground truth evidence sentences.

During training, we conduct message passing in $P(K,Y \, | \, X)$ (Section~\ref{sec:dpl}) iteratively, which usually converges within $5$ iterations. For distant supervision (Section~\ref{sec:noisy_gt}), we use the rule-based method to generate noisy labels for all the datasets except for RACE. On RACE, we use ILP-based method since we find the ILP-based method works better than the rule-based method on this dataset. The data programming and joint inference supervision on each dataset are slightly different. We will detail the differences in each subsection.

\subsubsection{Results on Multiple-Choice Datasets}
\label{sec:res:multiple}

\begin{figure}[!htbp]
   \centering
   \includegraphics[width=0.7\textwidth]{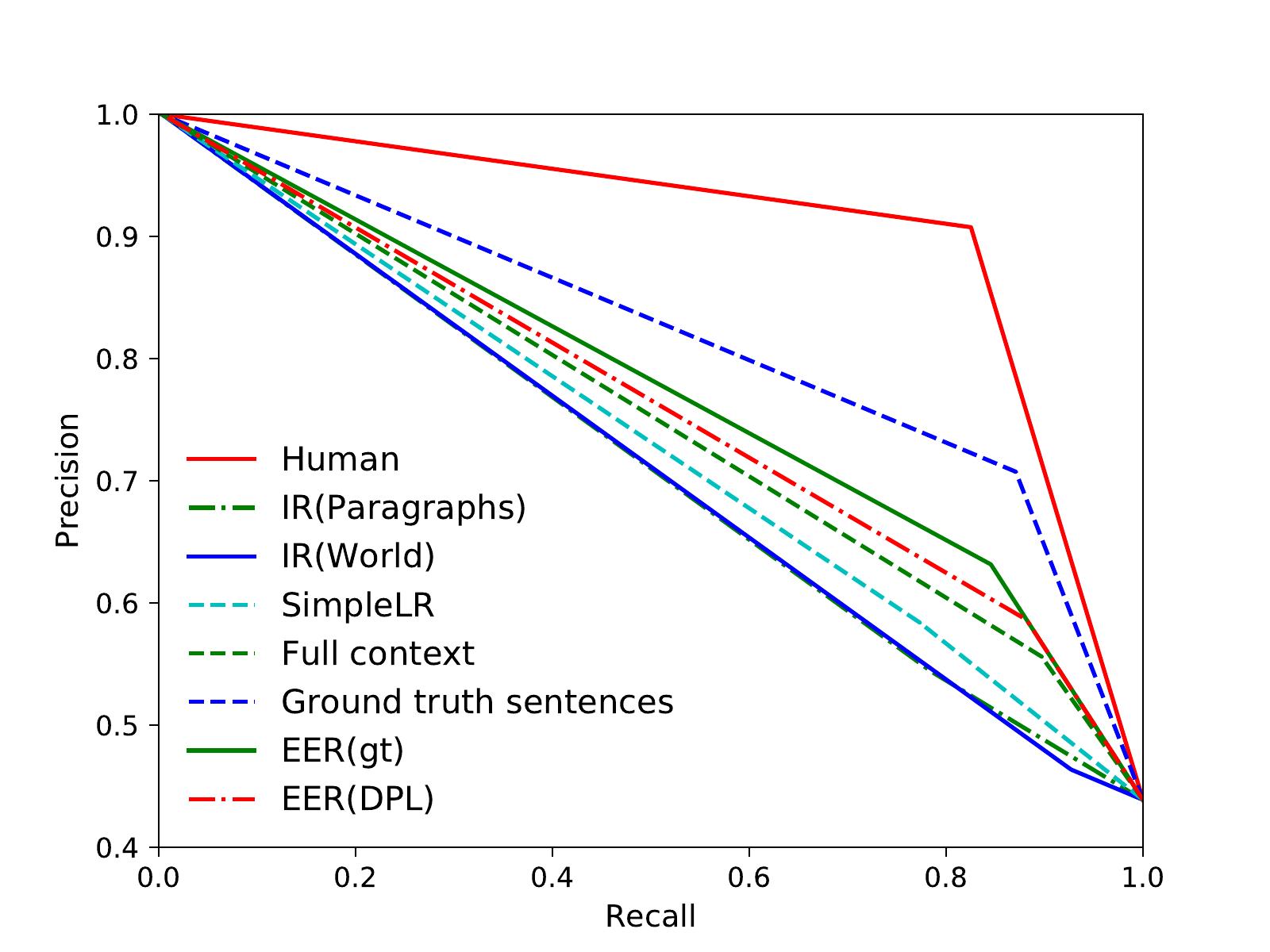}
 \caption{Precision-recall curves for different settings on the MultiRC development set (IR: information retrieval baseline; LR: logistic regression baseline implemented by~\citet{khashabi2018looking}).}
 \label{fig:multirc_result}
\end{figure}

\noindent \textbf{Evaluation on MultiRC:} Since its test set is not publicly available, currently we only evaluate our model on the development set (Table~\ref{tab:multirc_result}). Figure~\ref{fig:multirc_result} shows the precision-recall curves. The fine-tuned transformer (FT) baseline, which uses the full document as input, achieves an improvement of $2.2\%$ in macro-average F1 ($\text{F1}_{\text{m}}$) over the previous highest score, $66.5\%$. If we train our evidence extractor using the ground truth evidence sentences provided by turkers, we can obtain a much higher $\text{F1}_{\text{m}}$ $72.3\%$, even after we remove nearly $66\%$ of sentences in average per document. We can regard this result as the supervised upper bound for our evidence extractor. If we train the evidence extractor with KRDL as a supervision module, we get $70.5\%$ in $\text{F1}_{\text{m}}$. The performance gap between $70.5\%$ and $72.3\%$ shows there is still room for improving our denoising strategy.

\begin{table}[!htp]
\centering
\footnotesize
    \begin{tabular}{lccc}
    \toprule
     \textbf{Approach} & $\text{F1}_{m}$ & $\text{F1}_{a}$ & $\text{EM}_{0}$ \\ 
     \midrule
     All-ones baseline \cite{khashabi2018looking} & 61.0 & 59.9 & 0.8 \\
     Lucene world baseline \cite{khashabi2018looking} & 61.8 & 59.2 & 1.4 \\
     Lucene paragraphs baseline \cite{khashabi2018looking} & 64.3 &60.0 &7.5\\
     Logistic regression \cite{khashabi2018looking} & 66.5	& 63.2 & 11.8\\
     Full context + Fine-Tuned Transformer (FT,~\citet{radfordimproving}) & 68.7	& 66.7 & 11.0\\
     \midrule
     Random 5 sentences + FT  & 65.3	& 63.1 & 7.2	\\
     Top 5 sentences by $\text{EER}_{\text{DS}}~\text{+ FT}$  & 70.2 & \textbf{68.6} & 12.7 \\
     Top 5  sentences by $\text{EER}_{\text{KRDL}}~\text{+ FT}$ & \bf 70.5 & 67.8 & \bf 13.3 \\
     \midrule
     Top 5 sentences by $\text{EER}_{\text{gt}}~\text{+ FT}$ & \textbf{72.3} & \textbf{70.1} & \textbf{19.2} \\
     \midrule
     Ground truth evidence sentences + FT  & 78.1 & 74.0 & 28.6 \\
     Human Performance \cite{khashabi2018looking} & 86.4  & 83.8  &56.6 \\
     \bottomrule
  \end{tabular}
  \caption{Performance of various settings on the MultiRC development set. We use the same fine-tuned transformer (FT) as the evidence extractor (EER) and the neural reader ($\text{EER}_{\text{DS}}$: EER trained on the silver standard evidence sentences;
  $\text{EER}_{\text{KRDL}}$: EER trained with KRDL as a supervision module; $\text{EER}_{\text{gt}}$: EER trained using ground truth evidence sentences; $\text{F1}_\text{{m}}$\: macro-average F1; $\text{F1}_\text{{a}}$: micro-average F1; $\text{EM}_\text{{0}}$: exact match).
  }
  \label{tab:multirc_result}
 \end{table}

\noindent \textbf{Evaluation on DREAM:}
See Table~\ref{tab:dream_result} for results on DREAM dataset. The fine-tuned transformer (FT) baseline, which uses the full document as input, achieves test accuracy $55.1\%$. If we train our evidence extractor with KRDL as a supervision module and feed the extracted evidence sentences to the fine-tuned transformer, we get test accuracy $57.7\%$. Similarly, if we train the evidence extractor only with silver standard evidence sentences extracted from the rule-based distant supervision method, we obtain test accuracy $56.3\%$, \ie, $1.4\%$ lower than that with full supervision. Experiments demonstrate the effectiveness of our evidence extractor with denoising strategy, and the usefulness of evidence sentences for dialogue-based machine reading comprehension.

\begin{table}[h!t]
\footnotesize
\centering
  \begin{tabular}{lcc}
    \toprule
    \textbf{Approach} & \textbf{Dev} & \textbf{Test} \\ 
    \midrule
    Full context~\text{+ FT}$^{\dag}$~\cite{sundream2018} & 55.9 & 55.5 \\
    \midrule
    Full context~\text{+ FT}                     & 55.1 & 55.1 \\
    Top 3 sentences by $\text{EER}_{\text{silver-gt}}~\text{+ FT}$ & 50.1 & 50.4 \\
    Top 3 sentences by $\text{EER}_{\text{DS}}~\text{+ FT}$ & 55.1 & 56.3 \\
    Top 3 sentences by $\text{EER}_{\text{KRDL}}~\text{+ FT}$ & 57.3 & \textbf{57.7} \\
    \midrule
    Silver standard evidence sentences~\text{+ FT} & 60.5 & 59.8 \\
    Human Performance$^{\dag}$ & 93.9 & 95.5 \\
    \bottomrule
  \end{tabular}
\caption{Performance in accuracy (\%) on the DREAM dataset (Results marked with $^{\dag}$ are taken from~\citet{sundream2018}; $\text{EER}_{\text{silver-gt}}$: EER trained using silver standard evidence sentences).}
\label{tab:dream_result}
\end{table}

\noindent \textbf{Evaluation on RACE:}
On RACE, as we cannot find any public implementations of recently published independent sentence selectors, we compare our evidence sentence extractor with InferSent released by~\citet{conneau-EtAl} as previous work~\cite{htut2018training} has shown that it outperforms many state-of-the-art sophisticated sentence selectors on a range of tasks. We also investigate the \emph{\textbf{portability}} of our evidence extractor by combing it with two neural readers. Besides the fine-tuned transformer, we use Co-Matching~\cite{wang2018co}, another state-of-the-art neural reader on RACE.

As shown in Table~\ref{tab:race_result}, by using the evidence sentences selected by InferSent, we suffer up to a $1.9\%$ drop in accuracy with Co-Matching and up to a $4.2\%$ drop with the fine-tuned transformer. In comparison, by using the sentences extracted by our sentence extractor, which is trained with KRDL as a supervision module, we observe a much smaller decrease ($0.1\%$) in accuracy with the transformer baseline, and we slightly improve the accuracy with the Co-Matching baseline. For questions in RACE, introducing the content of answer options as additional information for evidence extraction can narrow the accuracy gap, which might be due to the fact that many questions are less informative~\cite{Yichong2018dynamic}. Note that all these results are compared with $59\%$ reported from ~\citet{radfordimproving}, if compared with our own replication ($56.8\%$), sentence extractor trained with either KRDL or distant supervision leads to gain up to $2.1\%$.

Since the problems in RACE are designed for human examinees that require advanced reading comprehension skills such as the utilization of external world knowledge and in-depth reasoning, even human annotators sometimes have difficulties in locating evidence sentences (Section~\ref{sec:human_eval}). Therefore, \emph{\textbf{a limited number of evidence sentences might be insufficient for answering challenging questions}}. Instead of removing ``non-relevant'' sentences, we keep all the sentences in a document while adding a special token before and after extracted evidence sentences. With KRDL as a supervision module, we see an improvement in accuracy of $0.9\%$ (from $58.9\%$ to $59.8\%$). 

For our current supervised upper bound (\ie, assuming we know the correct answer option, we find the silver evidence sentences from ILP-based distant supervision and then feed them into the fine-tuned transformer, we get $72.8\%$ in accuracy, which is quite close to the performance of Amazon Turkers. However, it is still much lower than the ceiling performance. To answer questions that require external knowledge, \emph{\textbf{it might be a promising direction to retrieve evidence sentences from external resources}}, compared to only considering sentences within a reference document.

\begin{center}
\begin{table}[!htbp]
\centering
\footnotesize
\begin{tabular}{lcccccc}
\toprule
     \multirow{2}{*}{\textbf{Approach}} & \multicolumn{3}{c}{\textbf{Dev}} & \multicolumn{3}{c}{\textbf{Test}}\\ 
     & Middle & High & All & Middle & High & All \\ 
     \midrule
     Sliding Window~\cite{richardson2013mctest,lai2017race} & - & - &-  & 37.3 & 30.4 & 32.2 \\ 
     Co-Matching~\cite{wang2018co} & - & - & - & 55.8 & 48.2 & 50.4 \\
     Full context + FT~\cite{radfordimproving} & - & - & - & 62.9 & 57.4 & 59.0 \\
     \midrule
     Full context + FT  & 55.6 & 56.5 & 56.0 & 57.5 & 56.5 & 56.8 \\
     Random 3 sents + FT & 50.3 & 51.1 & 50.9  & 50.9 & 49.5 & 49.9 \\ 
     \midrule
     Top 3 sents by InferSent (ques) + Co-Matching & 49.8 & 48.1 & 48.5 & 50.0 & 45.5 & 46.8 \\
     Top 3 sents by InferSent (ques + options) + Co-Matching & 52.6 & 49.2 & 50.1  & 52.6 & 46.8 & 48.5 \\ 
     Top 3 sents by $\text{EER}_{\text{DS}}$ + Co-Matching & 58.1 & 51.6 & 53.5 & 55.6 & 48.2 & 50.3 \\
     Top 3 sents by $\text{EER}_{\text{KRDL}}$ + Co-Matching & 57.5 & 52.9 & 54.2  & 57.5 & 49.3 & 51.6 \\
     \midrule
     Top 3 sents by InferSent (ques) + FT & 55.0 & 54.7 & 54.8  & 54.6 & 53.4 &   53.7 \\ 
     Top 3 sents by InferSent (ques + options) + FT & 59.2 & 54.6 & 55.9  & 57.2 & 53.8 & 54.8   \\ 
     Top 3 sents by $\text{EER}_{\text{DS}}$ + FT & 62.5 & 57.7 &  59.1 & 64.1 & 55.4 & 58.0 \\ 
     Top 3 sents by $\text{EER}_{\text{KRDL}}$ + FT & 63.2 & 56.9 & 58.8 & \textbf{64.3} & 56.7 & 58.9 \\
     \midrule
     Top 3 sents by $\text{EER}_{\text{DS}}$ + full context + FT & 63.4 & 58.6 & 60.0 & 63.7 & 57.7 & 59.5 \\
     Top 3 sents by $\text{EER}_{\text{KRDL}}$ + full context + FT & 64.2 & 58.5 & 60.2 & 62.4 & \textbf{58.7} & \textbf{59.8} \\ 
     \midrule
     Silver standard evidence sents + FT & 73.2 & 73.9 &  73.7 & 74.1  & 72.3 & 72.8  \\ 
     Amazon Turker Performance~\cite{lai2017race} & - & - & - & 85.1  & 69.4 & 73.3  \\ 
     Ceiling Performance~\cite{lai2017race} & - & - & - & 95.4 & 94.2 & 94.5  \\ 
     \bottomrule
  \end{tabular}
  \caption{Accuracy (\%) of various settings on the RACE dataset. $\text{EER}_{\text{DS}}$: evidence extractor trained on the silver standard evidence sentences extracted from the ILP-based distant supervision method.}
  \label{tab:race_result}
 \end{table}
\end{center}

\subsubsection{Human Evaluation}
\label{sec:human_eval}

\noindent \textbf{MultiRC}: Extracted evidence sentences, which help neural readers to find correct answers, may still fail to convince human readers. Thus we evaluate the quality of extracted evidence sentences based on human annotations (Table~\ref{tab:human_eval_result}). Even trained using the noisy labels, we achieve a macro-average F1 score $60.8\%$ on MultiRC, indicating the learning and generalization capabilities of our evidence extractor, compared to $53.0\%$, achieved by using the noisy silver standard evidence sentences guided by correct answer options.

\noindent \textbf{RACE}: Since RACE does not provide the ground truth evidence sentences, to get the ground truth evidence sentences, two internal annotators annotate $500$ questions from the RACE-Middle development set. The Cohen's kappa coefficient between two annotations is $0.87$. For negation questions which include negation words (\eg, \emph{Which statement is not true according to the passage?}), we have two annotation strategies: we can either find sentences that can directly imply the correct answer option; or the sentences that support the wrong answer options. During annotation, for each question, we use the strategy that leads to fewer evidence sentences. 

\emph{\textbf{We find that even humans have troubles in locating evidence sentences when the relationship between a question and its correct answer option is implicitly implied}}. For example, a significant number of questions require understanding the entire document (\eg, \emph{``what's the best title of this passage''} and \emph{``this passage mainly tells us that \_''}) and/or external knowledge (\eg, \emph{``the writer begins with the four questions in order to \_''}, \emph{``The passage is probably from \_''} , and \emph{``If the writer continues the article, he would most likely write about\_''}). For $10.8\%$ of total questions, at least one annotator leave the slot blank due to the challenges mentioned above. The average and the maximum number of evidence sentences for the remaining questions is $2.1$ and $8$ respectively. The average number of evidence sentences in whole RACE dataset should be higher since questions in RACE-High are more difficult~\cite{lai2017race}, and we ignore $10.8\%$ of the total questions which require understanding the whole context. In MultiRC, the average/maximum number of evidence sentences is $2.3$/$6$, respectively.

\begin{table}[!htbp]
\centering
\footnotesize
  \begin{tabular}{lcccccccccc}
  \toprule
  \multirow{1}{*}{\textbf{Dataset}}  & \multicolumn{3}{c}{\textbf{Quasar-T}} & \multicolumn{3}{c}{\textbf{SearchQA}} \\
  \midrule
 Model & Hits@1 & Hits@3 & Hits@5 & Hits@1 & Hits@3 & Hits@5 \\
\midrule
Information Retrieval~\cite{lin2018denoising}  & 6.3 &10.9 &15.2 &13.7& 24.1& 32.7  \\
INDEP~\cite{lin2018denoising}                 & 26.8 & 36.3 & 41.9  & 59.2 & 70.0 & 75.7  \\
FULL~\cite{lin2018denoising}                  & 27.7 & 36.8 & 42.6 & 58.9 & 69.8 & 75.5 \\
\midrule
$\text{EER}_{\text{KRDL}}$ & \textbf{42.3} & \textbf{56.7} & \textbf{62.0}  & \textbf{66.2} & \textbf{84.9} & \textbf{89.9}  \\
\bottomrule
\end{tabular}
\caption{Evidence extraction performance on two question answering datasets Quasar-T and SearchQA. INDEP: the sentence selector is trained independently; FULL: the sentence selector is trained jointly with a neural reader.}
\label{tab:span_result_hit}
\end{table}

\begin{table}[!htbp]
\centering
\begin{tabular}{lccc}
\toprule
\textbf{Dataset} & SE vs. GT & EER vs. SE & EER vs. GT \\ 
\midrule
RACE-M & 59.9 & 57.1  & 57.5  \\ 
MultiRC & 53.0 & 63.4 & 60.8 \\
\midrule
$\text{MultiRC}_{\text{gt}}$ & - & - & 63.1 \\
\bottomrule
\end{tabular}
\caption{Macro-average F1 with human annotations on the dev set (SE: silver standard evidence sentences; EER: evidence sentences extracted by EER trained on SE, GT: ground truth evidence sentences).}
\label{tab:human_eval_result}
\end{table}

\subsubsection{Results on Question Answering Datasets}

We are aware of some similar work~\cite{choi2017coarse,lin2018denoising,htut2018training} that aim to select relevant paragraphs for question answering tasks. Since most of them do not release implementations, we compare with ~\citet{lin2018denoising} on two open-domain question answering datasets since their work is most similar to ours and the code is available. We report a direct comparison between our evidence extractor and this state-of-the-art sentence selector in Table~\ref{tab:span_result_hit}. Our independently trained evidence extractor dramatically outperforms theirs, which is jointly trained with a neural reader. We obtain up to $52.7\%$ relative improvement on the Quasar-T dataset and $19.1\%$ relative improvement on the SearchQA dataset.

\subsection{Conclusions}

We propose an evidence extraction DNN trained with weak supervision. To denoise noisy labels, we combine various linguistic clues through knowledge rich deep learning. We equip state-of-the-art neural reader with extracted evidence sentences, and it achieves comparable or better performance than neural reader with full context on three datasets. Experimental results also show that our evidence sentence extractor is superior than other state-of-the-art sentence selectors. All those results indicate the effectiveness of our evidence extractor. For the future work, we aim to incorporate richer prior knowledge into KRDL, jointly train the evidence extraction DNN and neural readers, and create large-scale dataset that contains ground truth evidence sentences.

%% file: Chapters/multilingual.tex
\section{Improving Pre-Trained Multilingual Models with Vocabulary Expansion}

This chapter is based on our previous work ``Improving Pre-Trained Multilingual Model with Vocabulary Expansion"~\cite{wang2019improving}. It has been shown that performance on many natural language processing tasks drops dramatically on held-out data when a significant percentage of words do not appear in the training data, \ie, out-of-vocabulary (OOV) words~\cite{sogaard2012robust,madhyastha2016mapping}. A higher OOV rate (\ie, the percentage of the unseen words in the held-out data) may lead to a more severe performance drop~\cite{kaljahi2015foreebank}.
OOV problems have been addressed in previous works under monolingual settings, through replacing OOV words with their semantically similar in-vocabulary words~\cite{madhyastha2016mapping, kolachina2017replacing} or using character/word information~\cite{kim2016character,kim2018learning,chen2018combining} or subword information like byte pair encoding (BPE)~\cite{sennrich2016neural,stratos2017sub}. 

Recently, fine-tuning a pre-trained deep language model, such as Generative Pre-Training (GPT)~\cite{radford2018improving} and Bidirectional Encoder Representations from
Transformers (BERT)~\cite{devlin2018bert}, has achieved remarkable success on various downstream natural language processing tasks. Instead of pre-training many monolingual models like the existing English GPT, English BERT, and Chinese BERT, a more natural choice is to develop a powerful multilingual model such as the multilingual BERT.

However, all those pre-trained models rely on language modeling, where a common trick is to tie the weights of softmax and word embeddings~\cite{press2017using}. Due to the expensive computation of softmax~\cite{yang2017breaking} and data imbalance across different languages, the vocabulary size for each language in a multilingual model is relatively small compared to the monolingual BERT/GPT models, especially for low-resource languages. Even for a high-resource language like Chinese, its vocabulary size $10$k in the multilingual BERT is only half the size of that in the Chinese BERT. Just as in monolingual settings, the OOV problem also hinders the performance of a multilingual model on tasks that are sensitive to token-level or sentence-level information. For example, in the POS tagging problem (Table \ref{tab:pos_result}), 11 out of 16 languages have significant OOV issues (OOV rate $\ge 5\%$) when using multilingual BERT.


According to previous work~\cite{radford2018improving,devlin2018bert}, it is time-consuming and resource-intensive to pre-train a deep language model over large-scale corpora. To address the OOV problems, instead of pre-training a deep model with a large vocabulary, we aim at enlarging the vocabulary size when we fine-tune a pre-trained multilingual model on downstream tasks. 

We summarize our contributions as follows: (i) We investigate and compare two methods to alleviate the OOV issue. To the best of our knowledge, this is the first attempt to address the OOV problem in multilingual settings. (ii) By using English as an interlingua, we show that bilingual information helps alleviate the OOV issue, especially for low-resource languages. (iii) We conduct extensive experiments on a variety of token-level and sentence-level downstream tasks to examine the strengths and weaknesses of these methods, which may provide key insights into future directions.

\subsection{Related Work}

OOV poses challenges for many tasks~\cite{pinter2017mimicking} such as machine translation~\cite{razmara2013graph,sennrich2016neural} and sentiment analysis~\cite{kaewpitakkun2014sentiment}. Even for tasks such as machine reading comprehension that are less sensitive to the meanings of each word, OOV still hurts the performance~\cite{chu2017broad,zhang2018subword}. We now discuss previous methods in two settings.

\subsubsection{Monolingual Setting}

Most previous work address the OOV problems in monolingual settings. Before more fine-grained encoding schema such as BPE~\cite{sennrich2016neural} is proposed, prior work mainly focused on OOV for token-level representations~\cite{taylor2011towards,kolachina2017replacing}. Besides simply assigning random embeddings to unseen words~\cite{dhingra2017gated} or using an unique symbol to replace all these words with a shared embedding~\cite{hermann2015teaching}, a thread of research focuses on refining the OOV representations based on word-level information, such as using similar in-vocabulary words~\cite{tafforeau2015adapting,li2016towards,luong2015addressing,chousing2015}, mapping initial embedding to task-specific embedding~\cite{rothe2016ultradense,madhyastha2016mapping}, using definitions of OOV words from auxiliary data~\cite{long2016leveraging,bahdanau2017learning}, and tracking contexts to build/update representations~\cite{henaff2016tracking,kobayashi2017neural,ji2017dynamic,zhao2018addressing}.

Meanwhile, there have been efforts in representing words by utilizing character-level~\cite{zhang2015character,ling2015finding,ling2015character,kim2016character,gimpelcharagram2016} or subword-level representations~\cite{sennrich2016neural,bojanowski2017enriching}. To leverage the advantages in character and (sub)word level representation, some previous work combine (sub)word- and character-level representations~\cite{santos2014learning,dos2015boosting,yu2017joint} or develop hybrid word/subword-character architectures~\cite{chung2016character,luong2016achieving,pinter2017mimicking,li2018subword,bahdanau2017learning,matthews2018using}. However, all those approaches assume monolingual setting, which is different from ours.

\subsubsection{Multilingual Setting}
Addressing OOV problems in a multilingual setting is relatively under-explored, probably because most multilingual models use separate vocabularies~\cite{jaffe2017generating,platanios2018contextual}. While there is no direct precedent, previous work show that incorporating multilingual contexts can improve monolingual word embeddings~\cite{zou2013bilingual,ruder2017survey,andrew2013deep,faruqui2014improving,lu2015deep}.   

\citet{madhyastha2017learning} increase the vocabulary size for statistical machine translation (SMT). Given an OOV source word, they generate a translation list in target language, and integrate this list into SMT system. Although they also generate translation list (similar with us), their approach is still in monolingual setting with SMT. \citet{cotterell2017cross} train char-level taggers to predict morphological taggings for high/low resource languages jointly, alleviating OOV problems to some extent. In contrast, we focus on dealing with the OOV issue at subword level in the context of pre-trained BERT model.

\subsection{Approach}
\label{sec:method}

We use the multilingual BERT as the pre-trained model. We first introduce the pre-training procedure of this model (Section~\ref{sec:approach:bert}) and then introduce two methods we investigate to alleviate the OOV issue by expanding the vocabulary (Section~\ref{sec:approach:vocab}). Note that these approaches are not restricted to BERT but also applicable to other similar models, and these approaches can be seamlessly applied to alleviate both the sub-word and word level OOV issue.

\subsubsection{Pre-Trained BERT}
\label{sec:approach:bert}

Compared to GPT~\cite{radford2018improving} and ELMo~\cite{peters2018deep}, BERT~\cite{devlin2018bert} uses a \emph{\bf bidirectional} transformer, whereas GPT pre-trains a left-to-right transformer~\cite{liu2018generating}; ELMo~\cite{peters2018deep} independently trains left-to-right and right-to-left LSTMs~\cite{peters2017semi} to generate representations as additional features for end tasks. 

In the pre-training stage,~\citet{devlin2018bert} use two objectives: masked language model (LM) and next sentence prediction (NSP). In masked LM, they randomly mask some input tokens and then predict these masked tokens. Compared to unidirectional LM, masked LM enables representations to fuse the context from both directions. In the NSP task, given a certain sentence, it aims to predict the next sentence. The purpose of adding the NSP objective is that many downstream tasks such as question answering and language inference require sentence-level understanding, which is not directly captured by LM objectives.

\begin{figure*}[!htbp]
\begin{center}
\includegraphics[width=1.0\textwidth]{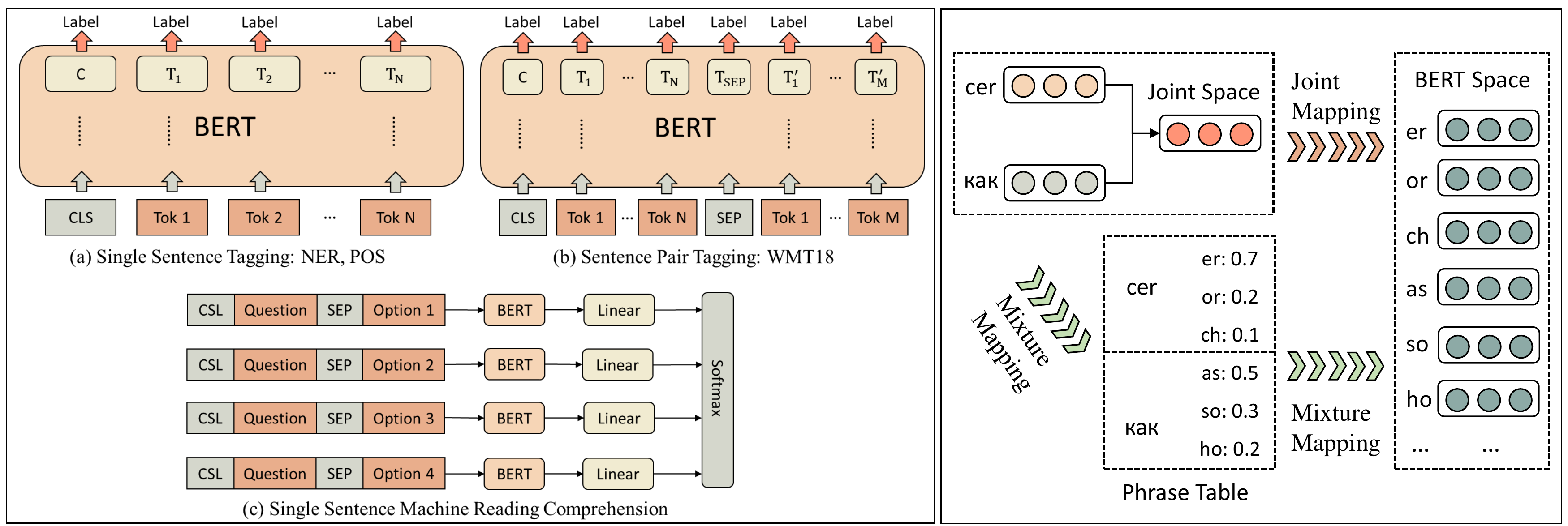}
\caption{Left: fine-tuning BERT on different kinds of end tasks. Right: illustration of joint and mixture mapping (in this example, during mixture mapping, we represent $\bm{e}(cer)=0.7*\bm{e}(er) + 0.2*\bm{e}(or) + 0.1*\bm{e}(ch)$).}
\label{fig:method:bertmerge}
\end{center}
\end{figure*}

After pre-training on large-scale corpora like Wikipedia and BookCorpus~\cite{zhu2015aligning}, we follow recent work~\cite{radford2018improving,devlin2018bert} to fine-tune the pre-trained model on different downstream tasks with minimal architecture adaptation. We show how to adapt BERT to different downstream tasks in Figure~\ref{fig:method:bertmerge} (left). 

\subsubsection{Vocabulary Expansion}
\label{sec:approach:vocab}


\citet{devlin2018bert} pre-train the multilingual BERT on Wikipedia in $102$ languages, with a shared vocabulary that contains $110$k subwords calculated from the WordPiece model~\cite{wu2016google}. If we ignore the shared subwords between languages, on average, each language has a $1.1$k vocabulary, which is significantly smaller than that of a monolingual pre-trained model such as GPT ($40$k). The OOV problem tends to be less serious for languages (\eg, French and Spanish) that belong to the same language family of English. However, this is not always true, especially for morphologically rich languages such as German~\cite{ataman2018compositional,lample2018phrase}. OOV problem is much more severe in low-resource scenarios, especially when a language (\eg, Japanese and Urdu) uses an entirely different character set from high-resource languages.

We focus on addressing the OOV issue at subword level in multilingual settings. Formally, suppose we have an embedding $E_{bert}$ extracted from the (non-contextualized) embedding layer in the multilingual BERT (i.e., the first layer of BERT). And suppose we have another set of (non-contextualized) sub-word embeddings $\{E_{l_{1}}, E_{l_{2}}, \dots, E_{l_{n}}\} \cup \{E_{en}\}$, which are pre-trained on large corpora using any standard word embedding toolkit. Specifically, $E_{en}$ represents the pre-trained embedding for English, and $E_{l_{i}}$ represents the pre-trained embedding for non-English language $l_{i}$ at the subword level. We denote the vocabulary of $E_{bert}$, $E_{en}$, and $E_{l_i}$ by $V_{bert}$, $V_{en}$, and $V_{l_{i}}$, respectively. For each subword $w$ in $V_{bert}$, we use $E_{bert}(w)$ to denote the pre-trained embedding of word $w$ in $E_{bert}$. $E_{l_i}(\cdot)$ and $E_{en}(\cdot)$ are defined in a similar way as $E_{bert}(\cdot)$. For each non-English language $l \in \{l_{1}, l_{2}, \dots, l_{n}\}$, we aim to enrich $E_{bert}$ with more subwords from the vocabulary in $E_{l_i}$ since $E_{l_i}$ contains a larger vocabulary of language $l_i$ compared to $E_{bert}$. 

As there is no previous work to address multilingual OOV issues, inspired by previous solutions designed for monolingual settings, we investigate the following two methods, and all of them can be applied at both word/subword level, though we find subword-level works better (Section~\ref{sec:experiment}). 


\noindent\textbf{Joint Mapping} For each non-English language $l$, we first construct a joint embedding space $E'_{l}$ through mapping $E_{l}$ to $E_{en}$ by an orthogonal mapping matrix $B_{l}$ (\ie, $E'_{l}=E_{l}B_{l}$). When a bilingual dictionary $f_l:V_{l} \to V_{en}$ is available or can be constructed based on the shared common subwords (Section~\ref{subsec:expsetting}), we obtain $B_{l}$ by minimizing:
\begin{equation}
\small
\sum_{w'\in V_{l} \cap \{ w: f_{l}(w) \in V_{en}\}} \!\!\!\!\!\!\!\!\!\! \norm{E_{l}(w')B_{l} - E_{en}(f_l(w'))}_{F}^{2} \nonumber
\end{equation}
where $\norm{\cdot}_{F}$ denotes the Frobenius norm. Otherwise, for language pair (\eg, English-Urdu) that meets neither of the above two conditions, we obtain $B_{l}$ by an unsupervised word alignment method from MUSE~\cite{conneau2017word}.

We then map $E'_{l}$ to $E_{bert}$ by an orthogonal mapping matrix $A'_{l}$, which is obtained by minimizing
\begin{equation}
\small
\sum_{w\in f_{l}(V_{l}) \cap V_{bert}} \norm{E'_{l}(w)A'_{l} - E_{bert}(w)}_{F}^{2} \nonumber
\end{equation}
We denote this method by $M_J$ in our discussion below, where the subscript $J$ stands for ``joint''.

\noindent\textbf{Mixture Mapping} Following the work of~\citet{gu2018universal} where they use English as \emph{``universal tokens''} and map all other languages to English to obtain the subword embeddings, we represent each subword in $E'_{l}$ (described in joint mapping) as a mixture of English subwords where those English subwords are already in the BERT vocab $V_{bert}$. This method, denoted by $M_{M}$, is also a joint mapping without the need for learning the mapping from $E'_{l}$ to $E_{bert}$. Specifically, for each $w\in V_{l}$, we obtain its embedding $\bm{e}(w)$ in the BERT embedding space $E_{bert}$ as follows. 
\begin{equation}
\small
 \bm{e}(w) = \sum_{u \in \mathcal{T}(w)} p(u|w)E_{bert}(u) \nonumber 
\end{equation}
where $\mathcal{T}(w)$ is a set to be defined later, and the mixture coefficient $p(u|w)$ is defined by
\begin{equation}
\small
    p(u|w) = \frac{\text{exp}(\text{CSLS}(E_{l}(u), E_{en}(w)))} {\sum_{v \in \mathcal{T}(w)}  \text{exp}(\text{CSLS}(E_{l}(v), E_{en}(w)))} \nonumber
\end{equation}%
where $\text{CSLS}$ refers to the distance metric Cross-domain Similarity Local Scaling~\cite{conneau2017word}. We select five $v \in V_{en}\cup V_{bert}$ with the highest $\text{CSLS}(E_{l}(v), E_{en}(w))$ to form set $\mathcal{T}(w)$. In all our experiments, we set the number of nearest neighbors in $\text{CSLS}$ to $10$. We refer readers to~\citet{conneau2017word} for details. Figure~\ref{fig:method:bertmerge} (right) illustrates the joint and mixture mapping.

\subsection{Experiment}
\label{sec:experiment}

\subsubsection{Experiment Settings}
\label{subsec:expsetting}

We obtain the pre-trained embeddings of a specific language by training fastText~\cite{bojanowski2017enriching} on Wikipedia articles in that language, with context window $5$ and negative sampling $5$. Before training, we first apply BPE~\cite{sennrich2016neural} to tokenize the corpus with subword vocabulary size $50$k. For joint mapping method $M_J$, we use bilingual dictionaries provided by \citet{conneau2017word}. For a language pair where a bilingual dictionary is not easily available, if two languages share a significant number of common subwords (this often happens when two languages belong to the same language family), we construct a bilingual dictionary based on the assumption that identical subwords have the same meaning~\cite{sogaard2018limitations}. We add all unseen subwords from $50$k vocabulary to BERT. We define a word as an OOV word once it cannot be represented as a single word. For example, in BERT, the sentence \emph{``Je sens qu' entre ça et les films de médecins et scientifiques"} is represented as \emph{``je sens qu \#\#' entre [UNK] et les films de [UNK] et scientifiques"}, where \textbf{qu'} is an OOV word since it can only be represented by two subword units: \textbf{qu} and \textbf{\#\#'}, but it is not OOV at subword level; \textbf{ça} and \textbf{médecins} cannot be represented by any single word or combination of subword units, and thus they are OOV at both word and subword level.    

We use the \textbf{multilingual BERT} with default parameters in all our experiments, except that we tune the batch size and training epochs. To have a thorough examination about the pros and cons of the explored methods, we conduct experiments on a variety of token-level and sentence-level classification tasks: part of speech (POS) tagging, named entity recognition (NER), machine translation quality estimation, and machine reading comprehension. See more details in each subsection.    

\subsubsection{Discussions about Mapping Methods}
Previous work typically assumes a linear mapping exists between embedding spaces of different languages if their embeddings are trained using similar techniques~\cite{xing2015normalized,madhyastha2016mapping}. However, it is difficult to map embeddings learned with different methods~\cite{sogaard2018limitations}. Considering the differences between BERT and fastText: \eg, the objectives, the way to differentiate between different subwords, and the much deeper architecture of BERT, it is very unlikely that the (non-contextualized) BERT embedding and fastText embedding reside in the same geometric space. Besides, due to that BERT has a relatively smaller vocabulary for each language, when we map a pre-trained vector to its portion in BERT indirectly as previous methods, the supervision signal is relatively weak, especially for low-resource languages. In our experiment, we find that the accuracy of the mapping from our pre-trained English embedding to multilingual BERT embedding (English portion) is lower than $30\%$. In contrast, the accuracy of the mapping between two regular English embeddings that are pre-trained using similar methods (e.g., CBOW or SkipGram~\cite{mikolov2013distributed}) could be above $95\%$~\cite{conneau2017word}.

Besides the joint mapping method $M_J$ (Section~\ref{sec:approach:vocab}), another possible method that could be used for OOV problem in multilingual setting is that, for each language $l$, we map its pre-trained embedding space $E_{l}$ to embedding $E_{bert}$ by an orthogonal mapping matrix $A_{l}$, which is obtained by minimizing $\sum_{w\in V_{l} \cap V_{bert}} \norm{E_{l}(w)A_{l} - E_{bert}(w)}_{F}^{2}$. This approach is similar to~\cite{madhyastha2016mapping}, and is referred as independent mapping method below. However, we use examples to demonstrate why these kind of methods are less promising. In Table~\ref{tab:result:mapping}, the first two rows are results obtained by mapping our pre-trained English embedding (using fastText) to the (non-contextualized) BERT embedding. In this new unified space, we align words with CSLS metric, and for each subword that appears in English Wikipedia, we seek its closest neighbor in the BERT vocabulary. Ideally, each word should find itself if it exists in the BERT vocabulary. However, this is not always true. For example, although \emph{``however"} exists in the BERT vocabulary, independent mapping fails to find it as its own closest neighbor. Instead, it incorrectly maps it to irrelevant Chinese words ``盘'' (\emph{``plate''}) and ``龙'' (\emph{``dragon''}). A similar phenomenon is observed for Chinese. For example, ``册'' is incorrectly aligned to Chinese words ``书'' and ``卷''.

\begin{table}[!ht]
\centering
\footnotesize
\begin{tabular}{llll}
\toprule
Source Lang & Source & Target & probability \\ 
\midrule
\multirow{2}{*}{English}& however & 盘 (plate)  & 0.91\\
&however & 龙 (dragon) & 0.05\\
\midrule
\multirow{2}{*}{Chinese}& 册 (booklet) & 书 (book) & 0.49 \\
&册 (booklet) & 卷 (volume) & 0.46 \\
\bottomrule
\end{tabular}
\caption{Alignment from Independent Mapping.}
\label{tab:result:mapping}
\end{table}

Furthermore, our POS tagging experiments (Section~\ref{sec:mono_seq}) further show that joint mapping $M_{J}$ does not improve (or even hurt) the performance of the multilingual BERT. Therefore, we use \textbf{mixture mapping} $M_{M}$ to address and discuss OOV issues in the remaining sections.


\begin{table}[!ht]
\centering
\footnotesize
  \begin{tabular}{l|ccc|cccc|cc}
    \toprule
    &
    \multicolumn{1}{c}{ $\text{BTS}^{\clubsuit}$ }  & \multicolumn{1}{c}{ $\text{BiLSTM}^{\diamondsuit}$ } & 
    \multicolumn{1}{c}{ $\text{FREQ}^{\diamondsuit}$ } &
    \multicolumn{1}{c}{\text{BERT}} & \multicolumn{1}{c}{$\text{BERT}_{\text{oov}}$} & \multicolumn{1}{c}{$\text{BERT}_{\text{oovR}}$} & 
    \multicolumn{1}{c}{$\text{BERT}_{\text{oovMJ}}$} &
    \multicolumn{1}{c}{ $\text{OOV}_{\text{w}}$} & \multicolumn{1}{c}{$\text{OOV}_{\text{sw}}$} \\
     \midrule
     ar &- & {98.23} & 90.06& 53.34 & \textbf{56.70} & 56.57  & 56.23 & 89.8 & 70.6 \\
     bg &97.84& 98.23& 90.06 & {\textbf{98.70}} & 98.22 & 94.41 & 97.21& 45.7 & 1.2 \\
     da &95.52 & 96.16& 96.35 & \textbf{97.16} & 96.53 & 94.15 &  94.85 & 38.9 & 2.8 \\
     de &92.87 & 93.51& 93.38 & 93.58 & \textbf{93.81} & 91.77  & 93.12 & 43.2 & 5.6 \\
     es &95.80 & 95.67& 95.74 & 96.04 & \textbf{96.92} & 95.10  & 95.77& 29.4 & 6.0\\
     fa &96.82 & {97.60} & 97.49  & 95.62 & 94.90 & 94.35  & \textbf{95.82}& 35.6 & 6.5 \\
     fi &95.48 &95.74& {95.85} & 87.72 & \textbf{93.35} & 84.75  & 89.39& 64.9 & 10.4 \\
     fr &95.75 & 96.20 &96.11 & 95.17 & \textbf{96.59} & 94.84  & 95.24& 33.9 & 10.3 \\
     hr &- & 96.27 & {96.82} & 95.03 & \textbf{96.49} & 89.87  & 93.48 & 49.5 & 8.3 \\
     it &97.56 &97.90 &97.95 & \textbf{98.22} & 98.00 & 97.63  & 97.85& 30.3 & 2.3 \\
     nl &- &92.82 &93.30 & \textbf{93.89} & 92.89 & 91.30  & 91.71& 35.5 & 0.3\\
     no &- & {98.06} &98.03 & \textbf{97.25} & 95.98 & 94.21  & 95.83 & 38.7 & 4.4 \\
     pl &- & {97.63} &97.62 & 91.62 & \textbf{95.95} & 87.50 & 92.56 & 56.5 & 13.6 \\
     pt &- & {97.94} &97.90 & 96.66 & \textbf{97.63} & 95.93 & 96.90 & 34.0 & 8.3 \\
     sl &- & {96.97} &96.84 & 95.02 & \textbf{96.91} & 89.55 & 93.97& 49.2 & 7.8 \\
     sv &95.57 &96.60 & {96.69} & 91.23 & \textbf{96.66} & 90.45 & 91.92 & 48.2 & 17.7 \\
\bottomrule
  \end{tabular}
  \caption{POS tagging accuracy (\%) on the Universal Dependencies v1.2 dataset. $\text{BERT}_{\text{oov}}$: BERT with method $M_{M}$. $\text{BERT}_{\text{oovR}}$: BERT with randomly picked embedding from BERT. $\text{BERT}_{\text{oovMJ}}$: BERT with method $M_{J}$. $\text{OOV}_{\text{w}}$: word-level OOV rate. $\text{OOV}_{\text{sw}}$: subword-level OOV rate. $\clubsuit$: \citet{gillick2016multilingual}, $\diamondsuit$: \citet{plank2016multilingual}.}
  \label{tab:pos_result}
 \end{table}

\begin{table}[!ht]
\centering
\footnotesize
  \begin{tabular}{l  ccc}
    \toprule
    \textbf{Approach}  & \bf Precision & \bf Recall & \bf F1 score \\
\midrule
DomainMask~\cite{peng2017multi}  & 60.8 & 44.9 & 51.7 \\
Linear Projection~\cite{peng2017multi} & 67.2 & 41.2 & 51.1 \\
Updates~\cite{peng2017supplementary}  & - & - & 56.1 \\
Updates~\cite{peng2017supplementary}  & - & -  & 59.0 \\
\midrule
BERT & 56.6 & 61.7 & 59.0 \\
$\text{BERT}_{\text{oov}} $ & 60.2 & 62.8 & \textbf{61.4} \\
$\text{BERT}_{\text{zh}} $ & 63.4 & 70.8 & \textbf{66.9} \\ 
\bottomrule
\end{tabular}
\caption{Performance of various models on the test set of Weibo NER. $\text{BERT}_{\text{zh}}$: Chinese BERT pre-trained over Chinese Wikipedia. We use scripts \textit{conlleval} for evaluation on NER.}
\label{tab:weiboner_result}
\end{table}

\subsubsection{Monolingual Sequence Labeling Tasks}
\label{sec:mono_seq}
\textbf{POS Tagging}: We use the Universal Dependencies
v1.2 data~\cite{mcdonald2013universal}. For languages with token segmentation ambiguity, we use the gold segmentation following~\citet{plank2016multilingual}. We consider languages that have sufficient training data and filter out languages that have unsatisfying embedding alignments with English (accuracy is lower than $20.0\%$ measured by word alignment accuracy or $0.25$ by unsupervised metric in MUSE~\cite{conneau2017word}). Finally, we keep $16$ languages. We use the original multilingual BERT (without using CRF~\cite{Lafferty:2001:CRF:645530.655813} on top of it for sequence labeling) to tune hyperparameters on the dev set and use the fixed hyperparameters for the expanded multilingual model. We do not tune the parameters for each model separately. As shown in Table~\ref{tab:pos_result}, at both the word and subword level, the OOV rate in this dataset is quite high. Mixture mapping improves the accuracy on $10$ out of $16$ languages, leading to a $1.97\%$ absolute gain in average. We discuss the influence of alignments in Section~\ref{sec:discussion}.





\noindent\textbf{Chinese NER}: We are also interested in investigating the performance gap between the expanded multilingual model and a monolingual BERT that is pre-trained on a large-scale monolingual corpus. Currently, pre-trained monolingual BERT models are available in English and Chinese. As English has been used as the interlingua, we compare the expanded multilingual BERT and the Chinese BERT on a Chinese NER task, evaluated on the Weibo NER dataset constructed from social media by~\citet{peng2015named}. In the training set, the token-level OOV rate is $2.17\%$, and the subword-level OOV rate is $0.54\%$. We tune the hyperparameters of each model based on the development set separately and then use the best hyperparameters of each model for evaluation on the test set. 

As shown in Table~\ref{tab:weiboner_result}, the expanded model outperforms the multilingual BERT on the Weibo NER dataset. We boost the F1 score from $59.0\%$ to $61.4\%$. Compared to the Chinese BERT ($66.9\%$), there still exists a noticeable performance gap. One possible reason could be the grammatical differences between Chinese and English. As BERT uses the language model loss function for pre-training, the pre-trained Chinese BERT could better capture the language-specific information comapred to the multilingual BERT.

\subsubsection{Code-Mixed Sequence Labeling Tasks}
\label{sec:mixed}

As the multilingual BERT is pre-trained over $102$ languages, it should be able to handle code-mixed texts. Here we examine its performance and the effectiveness of the expanded model in mixed language scenarios, using two tasks as case studies. 

\noindent\textbf{Code-Switch Challenge}: We first evaluate on the CALCS-2018 code-switched task~\cite{calcs2018shtask}, which contains two NER tracks on Twitter social data: mixed English\&Spanish (en-es) and mixed Modern Standard Arabic\&Egyptian (ar-eg). Compared to traditional NER datasets constructed from news, the dataset contains a significant portion of uncommon tokens like hashtags and abbreviations, making it quite challenging. For example, in the en-es track, the token-level OOV rate is $44.6\%$, and the subword-level OOV rate is $3.1\%$; in the ar-eg track, the token-level OOV rate is $64.0\%$, and the subword-level OOV rate is $6.0\%$. As shown in Table~\ref{tab:mixed_result:twitter}, on ar-eg, we boost the F1 score from $74.7\%$ to $77.3\%$. However, we do not see similar gains on the en-es dataset, probably because that English and Spanish share a large number of subwords, and adding too many new subwords might prevent the model from utilizing the well pre-trained subwords embedding. See Section~\ref{sec:discussion} for more discussions.



\eat{
\cite{indra2018bilingual} code switch on en and es, method is simple and potentially we can use the data.
\cite{aguilar2018named}
train: 
en-es:
OOV@token: 0.446437
OOV@subword: 0.030605
en-es (enriched)
OOV@token: 0.416830
OOV@subword: 0.013943
train:
ar-et:
OOV@token: 0.64
OOV@subword: 0.06
ar-et(enriched)
OOV@token:0.54 
OOV@subword:0.00888
}

\begin{table}[!ht]
\centering
\footnotesize
\begin{tabular}{lcccccccc}
\toprule
    \multirow{0}{*}{} & \multicolumn{3}{c}{\textbf{en-es}} & \multicolumn{3}{c}{\textbf{ar-eg}} \\
    \textbf{Model} & Prec & Rec & F1 & Prec & Rec & F1 \\ 
    \midrule
    $\text{FAIR}^{\clubsuit}$ & - & - & 62.4 & - & - & 71.6 \\ 
    $\text{IIT}^{\clubsuit}$ & - & - & 63.8 & -& - & - \\
    \midrule
    $\text{FAIR}^{\diamondsuit}$ & - & - & 67.7 & - & - & \textbf{81.4} \\
    BERT & 72.7 & \textbf{63.6} &\textbf{67.8} & 73.8 & 75.6 & 74.7 \\ 
    $\text{BERT}_{\text{oov}}$ & \textbf{74.2} & 60.9 & 66.9 & \textbf{76.9} & \textbf{77.8} & \textbf{77.3}\\
    \bottomrule
  \end{tabular}
  \caption{Accuracy (\%) on the code-switch challenge. The top two rows are based on the test set, and the bottom three rows are based on the development set. $\clubsuit$: results from~\citet{calcs2018shtask}. $\diamondsuit$: results from~\citet{wang2018code}. }
  \label{tab:mixed_result:twitter}
 \end{table}

\noindent \textbf{Machine Translation Quality Estimation:}
 All previous experiments are based on well-curated data. Here we evaluate the expanded model on a language generation task, where sometimes the generated sentences are out-of-control. 

 We choose the automatic Machine Translation Quality Estimation task and use Task $2$ -- word-level quality estimation -- in WMT18~\cite{wmt2018}. Given a source sentence and its translation (\ie, target), this task aims to estimate the translation quality (\emph{``BAD"} or \emph{``OK"}) at each position: \eg, each token in the source and target sentence, each \textbf{gap} in the target sentence. We use English to German (en-de) SMT translation. On all three categories, the expanded model consistently outperforms the original multilingual BERT (Table~\ref{tab:mixed_result:wmt})\footnote{Our evaluation is based on the development set since the test set is only available to participants, and we could not find the submission teams' performance on the developmenet set.}.
 
\begin{table}[!ht]
\centering
\footnotesize
\begin{tabular}{l ccc ccc ccc}
\toprule
    \multirow{0}{*}{} & \multicolumn{3}{c}{\textbf{Words in MT}} & \multicolumn{3}{c}{\textbf{Gaps in MT}} &\multicolumn{3}{c}{\textbf{Words in SRC}} \\
    \textbf{Model} & F1-BAD & F1-OK & F1-multi & F1-BAD & F1-OK & F1-multi & F1-BAD & F1-OK & F1-multi \\ 
    \midrule
    \cite{fan2018bilingual} & 0.68 & 0.92 & \textbf{0.62} &- & - & -& -& - & -   \\ 
    \cite{fan2018bilingual} & 0.66 & 0.92 & 0.61 & 0.51 & 0.98 & \textbf{0.50} &-  &- &- \\ 
    $\text{SHEF}-\text{PT}^{\clubsuit}$ & 0.51 & 0.85 & 0.43 & 0.29 & 0.96 & 0.28 & 0.42 & 0.80 & 0.34 \\
    \midrule
    BERT & 0.58 & 0.91 & 0.53 & 0.47 & 0.98 & 0.46 & 0.48 & 0.90 & 0.43   \\ 
    $\text{BERT}_{\text{oov}}$ & 0.60 & 0.91 & \textbf{0.55} & 0.50 & 0.98 & \textbf{0.49} & 0.49 & 0.90 & \textbf{0.44}   \\ 
    \bottomrule
  \end{tabular}
  \caption{WMT18 Quality Estimation Task 2 for the en$\to$de SMT dataset. $\clubsuit$: result from \citet{specia2018findings}. \textbf{MT}: machine translation, \eg, target sentence, \textbf{SRC}: source sentence. F1-OK: F1 score for \emph{``OK"} class; F1-BAD: F1 score for \emph{``BAD"} class; F1-multi: multiplication of F1-OK and F1-BAD.}
  \label{tab:mixed_result:wmt}
 \end{table}

\subsubsection{Sequence Classification Tasks}
\label{sec:xnli}
 
Finally, we evaluate the expanded model on sequence classification in a mixed-code setting, where results are less sensitive to unseen words.  

\noindent\textbf{Code-Mixed Machine Reading Comprehension}:
We consider the mixed-language machine reading comprehension task. Since there is no such public available dataset, we construct a new Chinese-English code-mixed machine reading comprehension dataset based on 37,436 unduplicated utterances obtained from the transcriptions of a Chinese and English mixed speech recognition corpus King-ASR-065-1\footnote{http://kingline.speechocean.com.}. We generate a multiple-choice machine reading comprehension problem (\ie, a question and four answer options) for each utterance. A question is an utterance with an English text span removed (we randomly pick one if there are multiple English spans) and the correct answer option is the removed English span. Distractors (\ie, wrong answer options) come from the top three closest English text spans, which appear in the corpus, based on the cosine similarity of word embeddings trained on the same corpus. For example, given a question ``突然听到~21\underline{~~~~~}，那强劲的鼓点，那一张张脸。'' (\emph{``Suddenly I heard 21\underline{~~~~~}, and the powerful drum beats reminded me of the players.''}) and four answer options \{ \emph{``forever''}, \emph{``guns''}, \emph{``jay''}, \emph{``twins'' }\}, the task is to select the correct answer option \emph{``guns''} (\emph{``21 Guns''} is a song by the American rock band \emph{Green Day}). We split the dataset into training, development, and testing of size 36,636, 400, 400, respectively.\footnote{We will release the code/annotations upon publication.} Annotators manually clean and improve the quality problems by generating more confusing distractors in the dev and testing sets to guarantee that these problems are error-free and challenging. 

In this experiment, for each BERT model, we follow its default hyperparameters. As shown in Table~\ref{tab:mixed_result:RC}, the expanded model improves the multilingual BERT ($38.1\%$) by $1.2\%$ in accuracy. Human performance ($81.4\%$) indicates that this is not an easy task even for human readers. 


\begin{table}[!ht]
\centering
\footnotesize
\begin{tabular}{lcc}
\toprule
    \multirow{0}{*}{} & \multicolumn{2}{c}{\textbf{Accuracy}} \\
    \textbf{Model} & Development & Test \\ 
     \midrule
    $\text{BERT}_{\text{en}}$ & 38.2 & 37.3   \\ 
    BERT & 38.7 & 38.1   \\ 
    \midrule
    $\text{BERT}_{\text{oov}}$ & 39.4 & \textbf{39.3} \\
    $\text{BERT}_{\text{zh}}$ & 40.0 & \textbf{45.0} \\
    \bottomrule
  \end{tabular}
  \caption{Accuracy (\%) of models on the code-mixed reading comprehension dataset. $\text{BERT}_{\text{en}}$: pre-trained English BERT. $\text{BERT}_{\text{zh}}$: pre-trained Chinese BERT.}
  \label{tab:mixed_result:RC}
 \end{table}


\subsubsection{Discussions}
\label{sec:discussion}

In this section, we first briefly investigate whether the performance boost indeed comes from the reduction of OOV and then discuss the strengths and weaknesses of the methods we investigate. 

First, we argue that it is essential to alleviate the OOV issue in multilingual settings. Taking the POS tagging task as an example, we find that most errors occur at the OOV positions (Table~\ref{tab:pos_result:analysis} in Section~\ref{sec:mono_seq}). In the original BERT, the accuracy of OOV words is much lower than that of non-OOV words, and we significantly boost the accuracy of OOV words with the expanded BERT. All these results indicate that the overall improvement mostly comes from the reduction of OOV.

\begin{table}[!ht]
\centering
\footnotesize
\begin{tabular}{lcccc}
\toprule
    \multirow{0}{*}{} & \multicolumn{2}{c}{BERT} & \multicolumn{2}{c}{$\text{BERT}_{\text{oov}}$} \\
    \textbf{Lang} & non-OOV & OOV & non-OOV & OOV \\ 
    \midrule
     fi & 98.1 & 81.3 & 98.5 & 90.2   \\
     fr & 97.0 & 90.2 & 97.2 & 95.6   \\
     hr & 97.8 & 91.9 & 97.7 & 94.5   \\
     pl & 98.8 & 84.6 & 99.0 & 93.2   \\
     pt & 98.8 & 91.5 & 98.6 & 94.8   \\
     sl & 98.6 & 91.6 & 98.7 & 95.1   \\
     sv & 97.4 & 82.9 & 98.2 & 94.8    \\
  \bottomrule
  \end{tabular}
  \caption{POS tagging accuracy (\%) for OOV tokens and non-OOV tokens on the Universal Dependencies v1.2 dataset, where the OOV/non-OOV are defined at word level with the original BERT vocabulary.}
  \label{tab:pos_result:analysis}
 \end{table}
 

We also observe that the following factors may influence the performance of the expanded model.


\noindent\textbf{Subwords}: When expanding the vocabulary, it is critical to add only frequent subwords. Currently, we add all unseen subwords from the $50$k vocabulary (Section \ref{subsec:expsetting}), which may be not an optimal choice. Adding too many subwords may prevent the model from utilizing the information from pre-trained subword embedding in BERT, especially when there is a low word-level overlap between the training and test set. 

\noindent\textbf{Language}: We also find that languages can influence the performance of the vocabulary expansion through the following two aspects: the alignment accuracy and the closeness between a language and English. For languages that are closely related to English such as French and Dutch, it is relatively easy to align their embeddings to English as most subword units are shared~\cite{sogaard2018limitations,conneau2017word}. In such case, the BERT embedding already contains sufficient information, and therefore adding additional subwords may hurt the performance. On the other hand, for a distant language such as Polish (Slavic family), which shares some subwords with English (Germanic family), adding subwords to BERT brings performance improvements. In the meantime, as Slavic and Germanic are two subdivisions of the Indo-European languages, we find that the embedding alignment methods perform reasonably well. For these languages, vocabulary expansion is usually more effective, indicated by POS tagging accuracies for Polish, Portuguese, and Slovenian (Table~\ref{tab:pos_result}). For more distant languages like Arabic (Semitic family) that use different character sets, it is necessary to add additional subwords. However, as the grammar of such a language is very different from that of English, how to accurately align their embeddings is the main bottleneck. 

\noindent\textbf{Task}: We see more significant performance gains on NER, POS and MT Quality Estimation, possibly because token-level understanding is more critical for these tasks, therefore alleviating OOV helps more. In comparison, for sequence level classification tasks such as machine reading comprehension (Section~\ref{sec:xnli}), OOV issue is less severe since the result is based on the entire sentence.

\subsection{Conclusion}

We investigated two methods (\ie, joint mapping and mixture mapping) inspired by monolingual solutions to alleviate the OOV issue in multilingual settings. Experimental results on several benchmarks demonstrate the effectiveness of mixture mapping and the usefulness of bilingual information. To the best of our knowledge, this is the first work to address and discuss OOV issues at the subword level in multilingual settings. Future work includes: investigating other embedding alignment methods such as Gromov-Wasserstein alignment~\cite{alvarez2018gromov} upon more languages; investigating approaches to choose the subwords to be added dynamically.

%% file: Chapters/episodic_memory.tex
\section{Large Episodic Memory Language Modelling}

This chapter is based on our previous work ``On-The-Fly Information Retrieval Augmentation for Language Models"~\cite{wang2020fly}. Here, we are interested in exploring the value of long term episodic memory in language modeling. For example, a language model can be used in January to assign a probability distribution over the statements that will appear in the newspaper in March.  But one month later, in February, the distribution over the predictions for March should be updated to take into account factual developments since the previous prediction. Long term episodic memory should be taken into account when assigning a probability to a statement.

Here we take a simple approach in which a pre-trained GPT language model~\cite{radford2018improving,radford2019language} is zero-shot augmented with an episodic memory consisting simply of a corpus of past news articles. Conceptually the past news articles are viewed as additional training data
which can be legitimately accessed when evaluating on future text. In our most basic experiment we calculate the probability of a future article by first calculating the probability of its first $k$ sentences using the pre-trained GPT model. We then use the first $k$ sentences as a query in an information retrieval system to extract a relevant past article. We then insert the past article following the first $k$ sentences when calculating the probability of the remainder of the future article using the same pre-trained GPT model. This is a zero-shot augmentation in the sense that there is no additional training or fine tuning of the pre-trained model. Our results show that this augmentation significantly reduces perplexity. We also present various other experiments including results on fine-tuning the model in the presence of the memory and the effect of this memory on event co-reference.

\subsection{Related Work}

Various language models have utilized external knowledge or long contexts ~\cite{paperno2016lambada,yang2017leveraging,PengNiRo19,khandelwal2018sharp,ghosh2016contextual,lau2017topically,grave2016improving,parthasarathi2018extending}. But these papers do not address the question of whether additional context or external knowledge is useful as a zero-shot augmentation of large scale pre-trained NLP models.

The value of external knowledge has previously been demonstrated for NLP tasks such as natural language inference~\cite{chen2018neural,yang2019enhancing}, language generation~\cite{parthasarathi2018extending}, knowledge base completion~\cite{toutanova2015representing,das2017go} and question answering~\cite{sun2019pullnet,sun2018open,dhingra2017linguistic}. However, all those prior works assume the model is small and trained from scratch.

As large scale pre-trained models have become more powerful it is not immediately clear whether external resources can still add value. The only work we know of on using external resources in modern large scale models is~\citet{yang2019enhancing} where a human curated external lexical resource is used to improve BERT.

 Our approach bears some resemblance to
 neural cache models~\cite{grave2016improving}. However, neural cache models store past hidden states as memory and accesses them through a dot product with the current hidden states. This is different
 from retrieving knowledge from a corpus-sized
 memory.

\ignore{
Great success has been achieved by first pre-training deep neural networks and then fine-tuning them on downstream tasks~\cite{radford2018improving,devlin2018bert,radford2019language}. 
Instead of using LSTM~\cite{peters2018deep}, transformer~\cite{vaswani2017attention} has been a popular choice since they can capture long range structures. However, it's time-consuming and resource-intensive to pre-train a deep model such as GPT 2.0~\cite{radford2019language} and BERT~\cite{devlin2018bert}. It's also unclear whether the pre-trained models already learnt sufficient knowledge. 
}

Our approach is also somewhat related to memory networks~\cite{weston2014memory}. Memory networks have a memory module which can be learnt jointly with other components. It has shown success in applications such as machine reading comprehension~\cite{kumar2016ask,xiong2016dynamic,shi2016hierarchical} and visual question answering~\cite{na2017read,ma2018visual,su2018learning}. Significant progress in memory networks has been achieved in both architecture ~\cite{chandar2016hierarchical,miller2016key,gulcehre2017memory} and model scale~\cite{rae2016scaling, guillaume2019large}.

Several papers have formulated, and experimented with, scalable memory networks --- memory networks that employ some method of efficiently reading and writing to very large neural memories.  This is done with approximate nearest neighbor methods in~\citet{rae2016scaling} and with product keys in~\citet{guillaume2019large}. These large memories are used to provide additional model capacity where the memory contents are trained over a large data set
using gradient descent training, just as one would train the parameters of a very large network.  It is shown in~\citet{guillaume2019large} that it is possible to insert a large memory as a layer in a transformer architecture resulting a model where the same number of parameters and the same performance can be achieved with half the layers and with much faster training time than a standard transformer architecture.  Here, however, we are proposing zero-shot augmentation with an external data source used as an episodic memory.

The use of key-value memories in~\citet{miller2016key} is particularly similar to our model.  Key-value memories were used there in treating a corpus of Wikipedia movie pages as a memory for answering questions about movies. As in our system, articles were extracted using word based information retrieval. Each article was encoded as a vector which was then given to a question answering architecture. This was shown to improve on automated knowledge base extraction from the same corpus but was still not competitive with human curated knowledge graphs for movies. Here we give the text of the retrieved article directly to the language model architecture and focus on augmenting large scale language models.

\subsection{Model}
 
\begin{figure}[!htbp]
\centering
\includegraphics[scale=0.6]{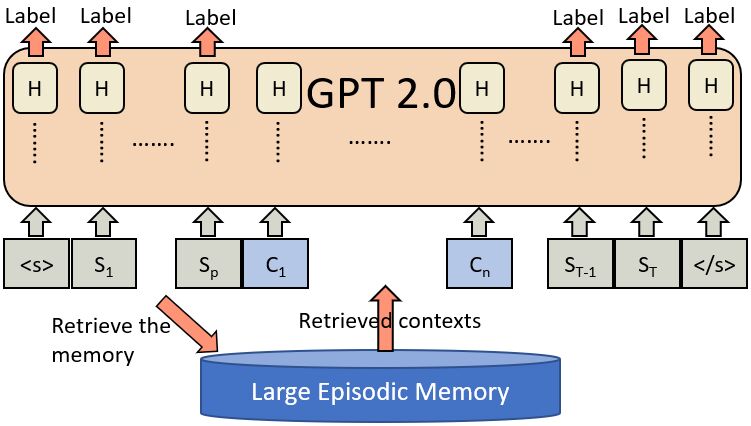}
\caption{GPT with large episodic memory component}
\label{fig:overview_model}
\end{figure}

We use the pre-trained transformer GPT 2.0~\cite{radford2019language}. Let $W_{w}$ and $W_{p}$ be the subword and position embeddings respectively. Let $M$ denote the total number of layers, for a token at time step $t$, the $m$-th layer's hidden state $h_{t}^{m}$ is given by:
\begin{equation}
\nonumber
h_{t}^{m} = \begin{cases}
W_{w} + W_{p} &\text{if $m=0$}\\
\text{TB}(h_{t}^{m-1}) & \text{if $1\leq m \leq M$}
\end{cases}
\end{equation}

where TB stands for Transformer Block, and it containing a MLP, residual connection~\cite{he2016deep}, self attention~\cite{vaswani2017attention} and LayerNorm~\cite{ba2016layer}. 
We use last layer's hidden state $h_{t}^{M}$ as the presentation $H_{t}$ for the token at time step $t$. We augment GPT 2.0 with a large episodic memory component, and the overall architecture is shown in Figure \ref{fig:overview_model}. 

For a sequence $S$ with $T$ tokens, let $S_1$, $\ldots$, $S_p$ be the tokens of the first $k$ sentences. Let $C$ be a sequence (article) retrieved from memory using the first $k$ sentences as the query, the vector $H_{t}$ is:
\begin{equation}
\nonumber
H_{t} = \begin{cases}
\text{GPT}(S_{1},\dots, S_{t}),\text{if $t \leq p$}& \\
\text{GPT}(S_{1},\dots,S_p,C,\ldots, S_{t}),\text{otherwise}&
\end{cases}
\end{equation}

That's to say, for the first $k$ sentences, we directly feed them to GPT to obtain their representations. For remaining sentences, their representations are conditioned on both the first $k$ sentences and the retrieved context $C$. Table \ref{tab:comparision} compares features of our simple memory augmentation with those of other memory models.

\begin{table}[!htbp]
\begin{center}
\begin{tabular}{lccc}
\bf Model & \bf episodic & \bf search & \bf memory size \\
\toprule
 DMN & yes  & exact & $\sim$1K words \\
 SAM: & no & approx  & $\sim$100K slots \\
 KVM: & yes  & exact &  $\leq$ 1M slots \\
 LMN: & no  & exact & $\sim$1M slots \\
 Ours: & yes  & approx & $\sim$10M documents \\
\end{tabular}
\caption{Comparison between different models. DMN: Dynamic Memory Network~\cite{xiong2016dynamic}; SAM: Sparse Access Memory~\cite{rae2016scaling}; KVM: Key Value Memory~\cite{miller2016key}; LMN: Large Memory Network~\cite{guillaume2019large}. Memory size is measured in their own words.}
\label{tab:comparision}
\end{center}
\end{table}

\subsection{Experiments}

 We focus on two tasks: document level language modelling and event co-retrieved . In both tasks we take a document as input and use first $k$ sentences to query the memory. To calculate the perplexity of a document, we compute the log-probability of a document by multiplying byte level probability, then divide the log-probability by the actual word count in the \textit{query} document. 
 
We use Gigaword~\cite{parker2011english} as both our language modeling
test set and as our external memory. Gigaword contains news from different sources such as NY Times and XinHua News etc. For language modelling we use the NY Times portion because it is written by native English speakers. Since GPT 2.0 is trained on Common Crawl
which contains news collections started from 2008. To avoid testing on GPT-2 training data, we use Gigaword articles collected prior to 2008. For the pre-trained language model we use GPT 2.0~\cite{radford2019language} \footnote{https://github.com/huggingface/pytorch-transformers}. It contains three pre-trained models: GPT Small, Medium and Large.

For information retrieval we use Lucene due to its simplicity. 
Given a query document we first do sentence and word tokenization and then use the first $k$ sentences to retrieve top 20 retrieved documents with the default TF-IDF distance metric provided by Lucene. Since too distant document pairs are uninformative and too related document pairs tends to be duplicates of the test article, we further filter those top ranked documents by time stamp, news source and cosine similarity. More specifically, we choose the highest ranked retrieved  document that simultaneously satisfies the following three conditions: it comes from a different news source; it appears earlier but within two weeks time window of the test document, and the bag of word cosine similarity between the test and the retrieved  cannot be larger than $0.6\alpha$ where $\alpha$ is the largest bag of word cosine similarity between the test article and any retrieved articles. To support fine-tuning experiments we constructed
a corpus of pairs of a \textit{query} article and a cached \textit{retrieved} document. We split the dataset into train/dev/test by query document's time stamp. The train/dev/test size is: 79622,16927,8045. For zero-shot experiments we use the test set of 8045 articles.  We do experiments with $k \in \{1,2,5\}$.

To check the quality of \textit{query-retrieved } pairs, we randomly sample 100 pairs from dev set and compute the bag of word cosine similarity between the two documents. The mean cosine similarity is 0.15. We also manually inspect them: we ask two NLP researchers to annotate the \textit{query-retrieved } pair as ``BAD" or ``OK" independently, i.e., if two documents are almost duplicates or totally unrelated, then it's ``BAD", otherwise, it's ``OK". Among 100 pairs, 83 pairs are ``OK", 17 pairs are ``BAD" due to irrelevance. The Cohen's kappa coefficient between two annotations is 0.94.

\subsubsection{Language modelling}
\label{exp:result:lm}

For language modeling we try zero-shot memory augmentation, fine-tuned memory augmentation, and training a small memory-augmented network from scratch. When training, we use the Adam optimizer from GPT 1.0~\cite{radfordimproving}. The learning rate is 0.001, weight decay parameter is 0.01, the warm up proportion is 0.1. For other parameters, we use the default values from GPT 2.0. The fine-tuning on Gigaword takes less than one day with a single GPU.

\smallpar{Zero-shot and fine-tuning results} Following ~\citet{radford2019language}, we first evaluate our model on Gigaword with zero-shot setting and then fine-tune the model. The results are given in Table \ref{result:zero-shot-giga}.  

\begin{table}[H]
\begin{center}
\begin{tabular}{lllll}
\multicolumn{1}{l}{\bf Model Size}  & \multicolumn{1}{c}{\bf woc} & \multicolumn{1}{c}{\bf k=1} & \multicolumn{1}{c}{\bf k=2} & \multicolumn{1}{c}{\bf k=5} \\
\midrule
GPT-Small         & 35.15 & 29.29 & 30.54 & 32.38 \\
GPT-Medium       & 22.78 & 19.84 & 20.54 & 21.48  \\
GPT-Large            & 19.90 & 17.41 & 18.00 & 18.80 \\
\midrule
GPT-Small         & 23.03 & 21.01 & 21.89 & 22.66 \\
\end{tabular}
\end{center}
\caption{Perplexity for zero-shot (top 3 rows) and fine-tuning (last row) settings when use different $k$ to retrieve the context. \textbf{woc}: without retrieved context.}
\label{result:zero-shot-giga}
\end{table}

From Table \ref{result:zero-shot-giga}, we see that with additional context retrieved from episodic memory, for all different GPT models, we obtain significantly lower perplexity than using original GPT 2.0. When fine tuning the model with context, we can further reduce the overall perplexity. We only fine tune GPT small due to our GPU memory constraints. Preliminary analysis indicates that most of the perplexity reduction comes at content words and semantically rich words where predictions require broader context. This is consistent with the phenomena found in~\citet{khandelwal2018sharp}. We further find that smaller $k$ leads to slightly worse retrieval quality, however, more continued sentences will benefit from the retrieved context. Since Gigaword contains newswire, the first several sentences usually are importation summarizations, thus overall, smaller $k$ will result in lower perplexity.     

\smallpar{Train from scratch} We also investigate training this form of memory-augmented model from scratch on our query-retrieved pairs. For these experiments we train smaller transformers and the results are given in Table \ref{result:train-giga}. From Table \ref{result:train-giga}, we see that additional context still helps and we can get decent perplexity even with quite small models.

\begin{table}[H]
\begin{center}
\begin{tabular}{lllll}
\multicolumn{1}{l}{\bf Model Config}  & \multicolumn{1}{c}{\bf woc} & \multicolumn{1}{c}{\bf k=1} & \multicolumn{1}{c}{\bf k=2} & \multicolumn{1}{c}{\bf k=5} \\
\midrule
E=384,H=6,L=6             & 35.62 & 31.94 & 33.18 & 35.26 \\
E=384,H=8,L=8             & 33.67 & 29.62 & 30.76 & 32.73 \\
E=576,H=8,L=8             & 31.32 & 27.38 & 28.54 & 30.63 \\
\end{tabular}
\end{center}
\caption{Perplexity when train from scratch. E: hidden states dimensionality; H: \# of head; L: \# of layer. GPT-Small has the configuration: E=764, H=12, L=12.}
\label{result:train-giga}
\end{table}

\smallpar{When context is less relevant} We also evaluate our method on Wikitext-2/103, in which the retrieved context is less relevant due to domain difference between Wikipedia and Gigaword. In this case, we use the most top ranked document from Gigaword as reference. Table \ref{result:zero-shot-wikitext} shows that less relevant contexts
have very little impact on perplexity.

\begin{table}[H]
\begin{center}
\begin{tabular}{lllll}
\multicolumn{1}{l}{\bf Dataset}  & \multicolumn{1}{c}{\bf woc} & \multicolumn{1}{c}{\bf k=1} & \multicolumn{1}{c}{\bf k=2} & \multicolumn{1}{c}{\bf k=5} \\
\midrule
Wikitext-2         & 28.67 & 28.96 & 28.95 & 28.70 \\
Wikitext-103         &25.38  & 25.68 & 25.56 & 25.39 \\
\end{tabular}
\end{center}
\caption{Zero-shot perplexity using GPT-Small}
\label{result:zero-shot-wikitext}
\end{table}

\subsubsection{Event Co-reference}

Intuitively episodic memory is useful because it contains information about the particular events mentioned in the test document. 
With this in mind we evaluate our approach on the event co-reference dataset ECB+~\cite{cybulska2014using}. ECB+ contains 982 documents clustered into 43 topics, and has two evaluation settings: coreferring mentions occurring within a single document (within document) or across a document collection (cross document). For the event co-reference pipeline, we follow the joint modeling method of~\citet{barhom2019revisiting} where they jointly represented entity and event mentions with various features and learned a pairwise mention/entity scorer for coreference classification. We augment their mention features with the mention's vector representations extracted from either GPT 2.0 or our zero-shot augmented GPT 2.0. For event co-reference, we use the whole test document to retrieve the context from Gigaword. From Table \ref{result:event-coref}, we see that the context can help boost the CONLL F1 score. 

\begin{table}[!htbp]
\begin{center}
\begin{tabular}{lccc}
\multicolumn{1}{l}{System} & \multicolumn{1}{c}{MUC} & \multicolumn{1}{c}{ $\text{B}^{3}$ } & \multicolumn{1}{c}{CONLL} \\
\toprule
\multicolumn{4}{l}{Within Document} \\
\midrule
KCP   & 63.0 & 92.0 & 81.0 \\
JM & 70.9 & 93.5 & 85.1 \\
JM+GPT & 80.1 & 93.5 & 85.2 \\
$\text{JM+GPT+CTX}^{\clubsuit}$ & 80.2 & 93.9 & 85.4 \\
\midrule
\multicolumn{4}{l}{Combined Within and Cross Document} \\
\midrule
CV   & 73.0 & 74.0 & 73.0 \\
KCP     & 69.0 & 69.0 &  69.0 \\
JM & 80.9 & 80.3 & 79.5 \\
JM+GPT & 81.2 & 80.2 & 79.6 \\
$\text{JM+GPT+CTX}^{\clubsuit}$ & 81.3 & 80.5 & 79.8 \\
\end{tabular}
\end{center}

\caption{F1 score on ECB+ dataset. KCP:~\citet{kenyon2018resolving} where they add a clustering-oriented regularization term;
CV:~\citet{cybulska2015translating} where they add the feature calculated from ``event template"; JM:~\citet{barhom2019revisiting}.
$\clubsuit$: we also feed the retrieved context to GPT to get the representation.}
\label{result:event-coref}
\end{table}

\subsection{Conclusion}

In this section, we propose a method to augment a pre-trained NLP model with a large episodic memory. Unlike previous work, we use information retrieval to handle a large external
corpus of text and feed retrieved documents directly to language models.
Evaluation results on language modelling and event co-reference show the promise of our method. To the best of our knowledge, this is the first work that augments pre-trained NLP models with large episodic memory. In principle, the memory-augmented GPT-2 can be used as a variant of GPT-2 for any downstream tasks, such as GLUE tasks~\cite{wang2018glue}, although we have not experimented with that here.

%% file: conclusion.tex
\section{Conclusion}

Deep learning is very powerful, but it's data hungry. Various approaches have been proposed to alleviate the annotation bottleneck for deep learning by making the model more knowledge efficient. In this thesis, we also propose several approaches to make the deep learning models more knowledge-efficient.

Specifically, we reviewed four work we have done in this direction: First, we proposed a knowledge rich deep learning model, which is an unifying learning framework for weak supervisions such as distant supervision, data programming and joint inference; Second, we applied knowledge rich deep learning model to assist the machine reading comprehension models to find the correct evidence sentences that can support their decision; Third, we investigate the knowledge transfer techniques in multilingual setting, where we proposed a method that can improve pre-trained multilingual BERT based on the bilingual dictionary; Last, we present an episodic memory network for language modelling, in which we encode the large external knowledge for the pre-trained GPT. 


We tried our best to make the deep learning models more knowledge-efficient, even the work we present in this thesis is not the most up-to-date, but they're my early exploration in this direction. Given the fact that current large scale unsupervised pre-training has began to revolutionize the NLP field, still, the power of unsupervised pre-training has not been fully discovered yet and lots of research problems remain unsolved such as efficient transfer learning and model compression etc. Most likely, our future work will center around large scale unsupervised pre-training.

%% file: Appendices/AppendixA.tex
\chapter{Publication List}

1: \textbf{Hai Wang}, David McAllester and Dan Roth, "On-The-Fly Information Retrieval Augmentation for Language Models", \textbf{ACL 2020 NUSE Workshop} \\
2: \textbf{Hai Wang}, Dian Yu, Kai Sun, Janshu Chen, Dong Yu, David McAllester and Dan Roth, "Evidence Extraction for Machine Reading Comprehension", \textbf{CONLL 2019} \\
3: \textbf{Hai Wang}, Dian Yu, Kai Sun, Janshu Chen, Dong Yu, "Improving Pre-Trained Multilingual Models with Vocabulary Expansion", \textbf{CONLL 2019} \\
4: \textbf{Hai Wang}, Hoifung Poon, "Deep Probabilistic Logic: A Unifying Framework for Indirect Supervision", \textbf{EMNLP 2018} \\
5: \textbf{Hai Wang*}, Takeshi Onishi\textbf{*}, Kevin Gimpel and David McAllester, "Emergent Predication Structure in Hidden State Vectors of Neural Readers", \textbf{ACL Workshop 2017} \\
6: Zewei Chu, \textbf{Hai Wang}, Kevin Gimpel and David McAllester, "Broad Context Language Modeling as Reading Comprehension", \textbf{EACL 2017} \\
6: Takaaki Hori, \textbf{Hai Wang}, Chiori Hori, et al, "Dialog State Tracking with Attention-based Sequence-to-Sequence Learning", \textbf{IEEE SLT 2016} \\
8: Takeshi Onishi, \textbf{Hai Wang}, Mohit Bansal, Kevin Gimpel and David McAllester, "Who did What: A Large-Scale Person-Centered Cloze Dataset", \textbf{EMNLP 2016} \\
9: \textbf{Hai Wang}, Mohit Bansal, Kevin Gimpel and David McAllester, "Machine Comprehension with Syntax, Frames, and Semantics", \textbf{ACL 2015} \\
10: Qixing Huang, \textbf{Hai Wang} and Vladlen Koltun, "Single-View Reconstruction via Joint Analysis of Image and Shape Collections", \textbf{Siggraph 2015} \\
11: Siqi Sun\textbf{*}, \textbf{Hai Wang*} and Jinbo Xu, "Inferring Block Structure of Graphical Models in Exponential Families", \textbf{AISTATS 2015}